\documentclass[10pt,journal]{IEEEtran}
\usepackage{bbm}
\usepackage{bm}

\usepackage{amsfonts}
\usepackage{amssymb}
\usepackage{algorithm}
\usepackage{algpseudocode}
\usepackage{graphicx}

\usepackage{subfigure}
\usepackage{color}
\usepackage{float}
\usepackage{stfloats}
\usepackage{longtable}
\usepackage{amsmath}
\usepackage{multirow}
\usepackage{url}
\newcommand{\trace}{trace}
\newcommand{\diag}{{diag}}

\ifCLASSINFOpdf
\else
\fi

\begin{document}

\title{Local Deep-Feature Alignment for Unsupervised Dimension Reduction}

\author{Jian~Zhang,
        Jun~Yu,~\IEEEmembership{Member,~IEEE,}
        and~Dacheng~Tao,~\IEEEmembership{Fellow,~IEEE}

\thanks{Manuscript received January 26, 2017; revised August 21, 2017 and
January 27, 2018; accepted February 4, 2018. This work was supported in
part by the Zhejiang Provincial Natural Science Foundation of China under
Grant LY17F020009, in part by the National Natural Science Foundation of
China under Grant 61303143, Grant 61622205, and Grant 61472110, and in
part by the Australian Research Council Projects under Grant FL-170100117,
Grant DP-180103424, Grant DP-140102164, and Grant LP-150100671. The
associate editor coordinating the review of this manuscript and approving it
for publication was Prof. Xiao-Ping Zhang. (Corresponding author: Jun Yu.)}

\thanks{J. Zhang is with the School of Science and Technology, Zhejiang International Studies University, Hangzhou 310012, China (email: jzhang@zisu.edu.cn).}
\thanks{J. Yu is with the School of Computer Science, Hangzhou Dianzi University, Hangzhou 310018, China (Corresponding author, email: yujun@hdu.edu.cn).}
\thanks{D. Tao is with the UBTECH Sydney Artificial Intelligence Centre and the School of Information Technologies, the Faculty of Engineering and Information Technologies, the University of Sydney, Darlington, NSW 2008, Australia (email: dacheng.tao@sydney.edu.au).}
}


\maketitle

\begin{abstract}
This paper presents an unsupervised deep-learning framework named
Local Deep-Feature Alignment (LDFA) for dimension reduction. We
construct neighbourhood for each data sample and learn a local
Stacked Contractive Auto-encoder (SCAE) from the neighbourhood to
extract the local deep features. Next, we exploit an affine
transformation to align the local deep features of each
neighbourhood with the global features. Moreover, we derive an
approach from LDFA to map explicitly a new data sample into the
learned low-dimensional subspace. The advantage of the LDFA method
is that it learns both local and global characteristics of the data
sample set: the local SCAEs capture local characteristics contained
in the data set, while the global alignment procedures encode the
interdependencies between neighbourhoods into the final
low-dimensional feature representations. Experimental results on data visualization,
clustering and classification show that the LDFA method is
competitive with several well-known dimension reduction techniques,
and exploiting locality in deep learning is a research topic worth
further exploring.
\end{abstract}


\IEEEpeerreviewmaketitle

\section{Introduction}

\IEEEPARstart{R}{ecent} years have witnessed an increase in the use
of deep learning in various research domains, such as audio
recognition, image and video analysis, and natural language
processing. Deep learning methods can be divided into two groups:
supervised learning and unsupervised learning. Supervised learning
aims to learn certain classification functions based on known
training samples and their labels for pattern recognition, while
unsupervised learning aims to learn useful representations from
unlabeled data. Both groups of methods have achieved great success,
but we are particularly interested in the unsupervised learning
methods because their mechanisms are closer to the learning
mechanism of human brain and simpler than those of supervised
methods \cite{lecun2015deep}.

A commonly used group of unsupervised deep-learning methods are
Auto-encoders (AEs). An AE learns a single transformation matrix for
embedding all the data, which means it does not discriminate between
data and treats every data sample in the same way. This is
coincident with some linear learning methods, such as PCA
\cite{pearson1901liii} and ICA \cite{hyvarinen2004independent},
which usually assume that the training data obey a single Gaussian
distribution. However, this assumption is not quite exact even for
the same kind of data. In fact, the data used in
various real-world applications often exhibit a trait of multiple
Gaussian distribution.
When we use multiple Gaussian models to depict the data
distribution, each Gaussian model in fact reflects some local
characteristic or the locality of the data set. Therefore it could
be important for the AEs to preserve the local characteristic during
feature learning. This concern has been supported by some previous
works \cite{alain2012regularized} in which an AE was equipped with a
regularization term, forcing it to be sensitive only to the data
variations along the data manifold where the locality was preserved.
However, the regularization term was originally designed for
improving robustness more than preserving locality.


A successful group of approaches to locality preservation are
manifold learning algorithms
\cite{roweis2000nonlinear,belkin2002laplacian,zhang2004principal,he2004locality}.
They exploit a structure called a neighbourhood graph to learn the
interrelations between data and transfer the interrelations to
low-dimensional space. Some algorithms assume that if some data are
close to one another in the high-dimensional space, they should also
be close in the low-dimensional space
\cite{belkin2002laplacian,he2004locality}. Some algorithms assume
that an unknown low-dimensional data sample can be reconstructed by
its neighbours in the same way as its high-dimensional counterpart
is reconstructed in the original space \cite{roweis2000nonlinear}.
These objectives differ largely from that of AEs. In comparison,
Local Tangent Space Alignment (LTSA) \cite{zhang2004principal}
computes a linear transformation for each neighbourhood to align the
local tangent-space coordinates of each neighbourhood with
the low-dimensional representations in a global
coordinate system. Since the local tangent-space coordinates can
exactly reconstruct each neighbourhood, LTSA shares more similarity
with AEs. The only difference between them is that each
neighbourhood in LTSA has a distinct reconstruction function and all
the data in AEs share the same reconstruction function.

Enlightened by LTSA, we propose an unsupervised deep-learning
framework for dimension reduction, in which the low-dimensional
feature representations are obtained by aligning local features of a
series of data subsets that capture the locality of the original
data set. Specifically, we construct a neighbourhood for each data
sample using the current sample and its neighbouring samples. Next,
we stack several regularized AEs (Contractive AE or CAE
\cite{alain2012regularized}) together to form a deep neural network
called Stacked CAE (SCAE) for mining local features from the
neighbourhood. We derive the final low-dimensional feature
representations by imposing a local affine transformation on the
features of each neighbourhood to transfer the features from each
local coordinate system to a global coordinate system. The local
features learned by each SCAE reflect the deep-level characteristics
of the neighbourhood, thus the proposed method can be named Local
Deep-Feature Alignment (LDFA). We also derive an explicit mapping
from the LDFA framework to map a new data sample to the learned
low-dimensional subspace. It is worthwhile to highlight several
aspects of the proposed method:

\begin{enumerate}[]

\item The locality characteristics contained in the neighbourhood can be
effectively preserved by the local SCAEs.

\item The regularization term of each SCAE facilitates estimating the parameters from a neighbourhood that usually does not contain
much data.

\item The number of "variations" of the local embedding function is small \cite{bengio2009learning} due to the data similarity among each neighbourhood, which reduces the difficulty in robust feature learning.

\item The local features are learned from a small amount of
data in a neighbourhood, so the proposed method can work well when
the data amount is not large.

\end{enumerate}

The rest of this paper is organized as follows: In Section 2, we
will review the related works to give the readers more insight into
deep learning and locality-preserving learning. In Section 3, we
will introduce the proposed method in detail. In Section 4, we will
show a series of experimental results on different applications.
Section 5 will present the paper's conclusions.

\section{Related works}
\label{review}

\subsection{Deep Learning}

Deep learning algorithms (DLA) receive much attention because they
can extract more representative features
\cite{jia2017sparse,li2016visual,liu2016robust} from data. The
current DLAs include supervised methods and unsupervised methods.
The most representative supervised methods are Convolutional Neural
Networks (CNNs), which are constructed by stacking three kinds of
layers together, i.e., the Convolutional Layer (CL), Pooling Layer
(PL), and Fully Connected Layer (FCL) \cite{cnn}. The CL and PL
differ from layers of regular neural networks in that the neurons in
each of these two layers are only connected to a
small region of the previous layer, instead of to all the neurons in
a fully connected manner. This greatly reduces the number of
parameters in the network and makes CNNs particularly suitable for
dealing with images. However, CNNs require every data sample to have
a label indicating the class tag of the sample, so they are not
applicable to unsupervised feature extraction.

The most representative unsupervised deep-learning methods are
AEs \cite{ap2014autoencoder}. An AE aims to learn the
low-dimensional feature representations that can best reconstruct
the original data.
Some researchers have enhanced AEs to increase the robustness to
noise \cite{alain2014regularized}. AEs are often used to construct
deep neural network structures for feature extraction
\cite{vincent2010stacked}.

However, traditional AEs train a single transformation matrix for
embedding all the data into low-dimensional space. Thus traditional
AEs capture only the global characteristics of the data and do not
consider the local characteristics. This might be inappropriate
because locality has proved to be a very useful characteristic in
pattern recognition \cite{liu2016unsupervised,tang2016local}.

\subsection{Locality-Preserving Learning}

Manifold learning methods are well known for their
capabilities of preserving the local characteristics of the data set
during dimension reduction
\cite{roweis2000nonlinear,belkin2002laplacian,zhang2004principal,he2004locality,roweis2002global}.
The local characteristics of the original data set are contained in
a structure called the neighbourhood graph, where each node
representing a data sample is connected to its nearest neighbuoring
nodes by arcs. The neighbourhood graph is then fed to some local
estimators \cite{bengio2009learning} that are capable of
transferring the locality to the learned low-dimensional feature
representations. Different manifold algorithms have different
locality-preservation strategies.

Locally Linear Embedding (LLE) \cite{roweis2000nonlinear}
reconstructs each data sample by linearly combining its neighboring
data samples, and assumes the low-dimensional feature representation
of the data sample can be reconstructed by its neighbours using the
same combination weights. Laplacian Eigenmap (LE)
\cite{belkin2002laplacian} assumes that if two data samples are
close to each other in the original data set, their low-dimensional
counterparts should also be close to each other. Locality Preserving
Projections (LPP) \cite{he2004locality} shares the same objective
with LE, but is realized in a linear way. Local Tangent Space
Alignment (LTSA) \cite{zhang2004principal} transfers the local
characteristics of the data set to a low-dimensional feature space
using a series of local Principal Component Analysis (PCA)
applications, and then obtains the low-dimensional feature
representations by aligning the local features learned by these
local PCAs. In LTSA, the local estimators realized by these local
PCAs are explicit and can be easily used for projecting data into
low-dimensional space. This is important in dimension reduction
algorithms. The locality characteristic of data can
also be preserved using the joint/conditional probability
distribution of data pairs based on their neighbouring structures. A
typical relevant method is t-distributed stochastic neighbor
embedding (t-SNE)\cite{maaten2008visualizing} that is particularly
effective in data visualization.

Some manifold-based methods encode discriminative
information in neighbourhood construction. The study in
\cite{su2014submanifold} simultaneously extracts a pair of manifolds
based on the data similarity among neighbourhoods such that the two
manifolds complement each other to enhance the discriminative power
of the features. In \cite{yan2009synchronized}, the neighbourhood of
each data is defined by the identity and pose information of a
subject, so that the learned manifolds can be applied to
person-independent human pose estimation.

However, Bengio confirmed that each of these manifold algorithms
could be reformed as a single-layer nonlinear neural network
\cite{bengio2009learning} that fails to discover deep-level features
from the original data. In addition, it does not make much sense to
stack manifold learning algorithms directly as a
layered structure to learn deep features because the objective
functions of manifold learning methods are
deliberately designed for one-layer learning. To achieve deep-level
feature learning based on manifold methods, we might want to combine
the locality-preservation capability of manifold methods with
deep-learning methods.

\subsection{Combination of Deep Learning and Locality Preserving Learning}

There is still no conclusion about how to encode the locality
learning into the deep-learning process. Rifai
\cite{alain2012regularized,rifai2011contractive} showed that a
certain kind of smoothness regularization might be useful in
preserving the data locality. In
\cite{alain2012regularized,rifai2011contractive}, the smoothness
regularization forces AEs to be sensitive only to the data
variations along the data manifold where locality is preserved.
However, the smoothness regularization was originally designed for
improving robustness more than preserving locality, so it only
indirectly models the data locality.

Some works have been proposed to enhance deep learning with some
straightforward locality-preserving constraints. A
Deep Adaptive Exemplar Auto-Encoder was proposed in
\cite{shao2016spectral} to extract deep discriminant features by
knowledge transferring. In this method, a low-rank coding
regularizer transfers the knowledge of the source domain to a shared
subspace with the target domain, while keeping the source and target
domains well aligned through the use of locality-awareness
reconstruction. This method is particularly useful in domain
adaptation where the data in source domain should be labeled.
\cite{lu2015multi} trained CNNs for each class of data and combined
the learning results of these CNNs at the top layers using the
maximal manifold-margin criterion. However, this criterion preserves
discriminative information between classes by directly evaluating
the Euclidean distances between the deep features learned by these
CNNs. This is not quite exact because the features
learned from different CNNs actually lie in
different coordinate systems. In addition, all the parameters of
these CNNs should be solved simultaneously in the feature-learning
process, which makes the optimization process highly nonlinear and
hard to solve. More importantly, the method can still not learn
features without data labels.

\section{Local Deep Feature Alignment Framework}
\label{method}

\subsection{Contractive Auto-Encoder (CAE)}\label{method1}
The original AEs are designed for learning a feature representation
from the input data sample that can be used to reconstruct the input
data sample as accurately as possible \cite{baldi2012autoencoders}.
Given a set of training data $\bm{{\rm X}} = [\bm{{\rm x}}_1,\bm{{\rm x}}_2,\dots,\bm{{\rm x}}_N] \in R^D$, the
encoding (embedding) and decoding (reconstruction) process can be
described as:
\begin{eqnarray}\label{enc_dec}
\left \{
\begin{aligned}
  & \bm{{\rm h}}_i = g(\bm{{\rm W x}}_i + \bm{{\rm b}}) \\
  & \bm{{\rm \tilde x}}_i = g(\bm{{\rm W}}^T \bm{{\rm h}}_i + \bm{{\rm c}}),
\end{aligned}
\right.
\end{eqnarray}
where $\bm{{\rm h}}_i \in R^d$ is the feature representation, $\bm{{\rm W}}$ describes the
weight matrix of the encoding, $\bm{{\rm b}}$ and $\bm{{\rm c}}$ are the bias terms,
$\bm{{\rm \tilde x}}_i$ is the reconstructed data sample and $g(\cdot)$
represents the activation function which is usually a sigmoid
function:
\begin{align}\label{sigmoid}
g(z) = \frac{1}{1 + exp(-z)}.
\end{align}

\begin{figure}[ht]
\centering
\includegraphics[width = 3.5in]{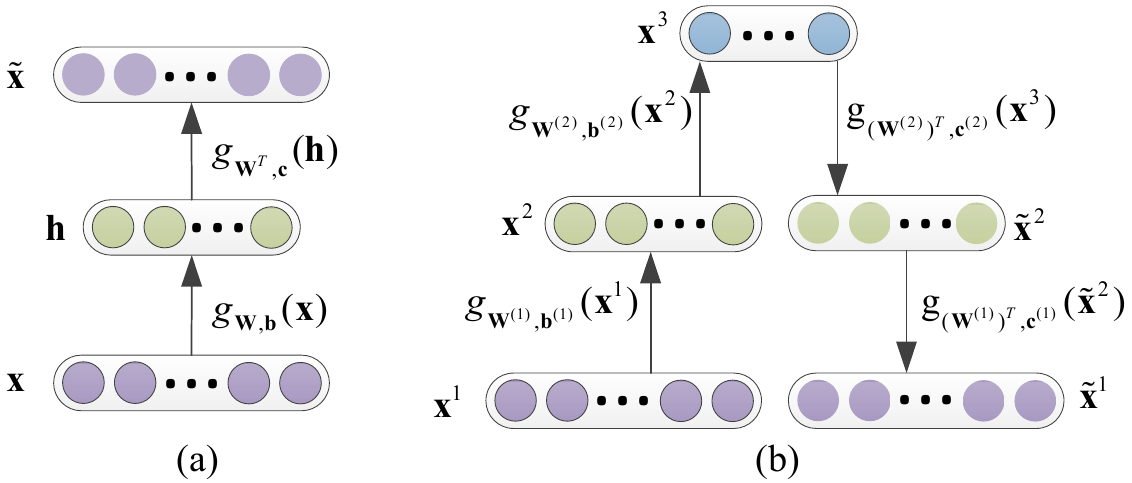}
\caption{ The Auto-encoder (AE) and a two-layer
Stacked Contractive Auto-encoder (SCAE). (a) The diagram of an AE
where $g_{\bm{{\rm W}},\bm{{\rm b}}}(\cdot)$ maps $\bm{{\rm x}}$ to
its feature representation $\bm{{\rm h}}$ that is used to
reconstruct $\bm{{\rm x}}$ as $\bm{{\rm \tilde x}}$ through
$g_{\bm{{\rm W}}^T,\bm{{\rm c}}}(\cdot)$. (b) The diagram of a
two-layer SCAE where $g_{\bm{{\rm W}}^{(l)},\bm{{\rm
b}}^{(l)}}(\cdot)(l=1,2)$ maps $\bm{{\rm x}}^1$ to $\bm{{\rm x}}^3$
that is used to obtain $\bm{{\rm \tilde x}}^1$ through $g_{(\bm{{\rm
W}}^{(l)})^T,\bm{{\rm c}}^{(l)}}(\cdot)(l=1,2)$. The superscript $l$
indicates the layer of the network.} \label{fig:sae}
\end{figure}

Next, the parameters of an AE can be obtained by solving the
following optimization problem:
\begin{align}\label{objae}
\min_{\bm{{\rm h}}_i,\bm{{\rm W}},\bm{{\rm b}},\bm{{\rm c}} } \sum_{i=1}^{N} \| \bm{{\rm h}}_i -g_{\bm{{\rm W}},\bm{{\rm b}}}(\bm{{\rm x}}_i)\|^2 + \|\bm{{\rm x}}_i -
g_{\bm{{\rm W}}^T,\bm{{\rm c}} }(\bm{{\rm h}}_i)\|^2.
\end{align}
The functionality of an AE is depicted in Fig. \ref{fig:sae} (a).

However, the learning process of an AE will be not robust enough in
some cases, for example, when the number of data samples is much
smaller than the dimension. To improve the robustness, some
researchers propose to use smoothness prior to
regularize the AE and thus derive the Contractive
Auto-encoder (CAE) \cite{rifai2011contractive} whose objective
function in matrix form is
\begin{align}\label{mxobjcae}
\begin{split}
&\| \bm{{\rm H}} -g_{\bm{{\rm W}},\bm{{\rm b}}}(\bm{{\rm X}})\|_F^2 + \|\bm{{\rm X}} - g_{\bm{{\rm W}}^T,\bm{{\rm c}}}(\bm{{\rm H}})\|_F^2 + \\
 &\lambda \trace \big ( \bm{{\rm W}}^T (\bm{{\rm H}} \odot (\bm{{\rm E}} - \bm{{\rm H}})) (\bm{{\rm H}} \odot (\bm{{\rm E}} - \bm{{\rm H}}))^T \bm{{\rm W}}
\big ),
\end{split}
\end{align}
where $\bm{{\rm H}} = [\bm{{\rm h}}_1,\dots,\bm{{\rm h}}_N]$,
$\bm{{\rm E}}$ is a matrix in which every entry is equal to one, and
$\odot$ denotes the dot product. The third term in (\ref{mxobjcae})
(the smoothness regularization term)
keeps the feature learning process insensitive to
data variations while competent for data reconstruction. This helps
to extract robust low-dimensional features
\cite{rifai2011contractive}.



\begin{figure*}
\centering
\includegraphics[width = 5in]{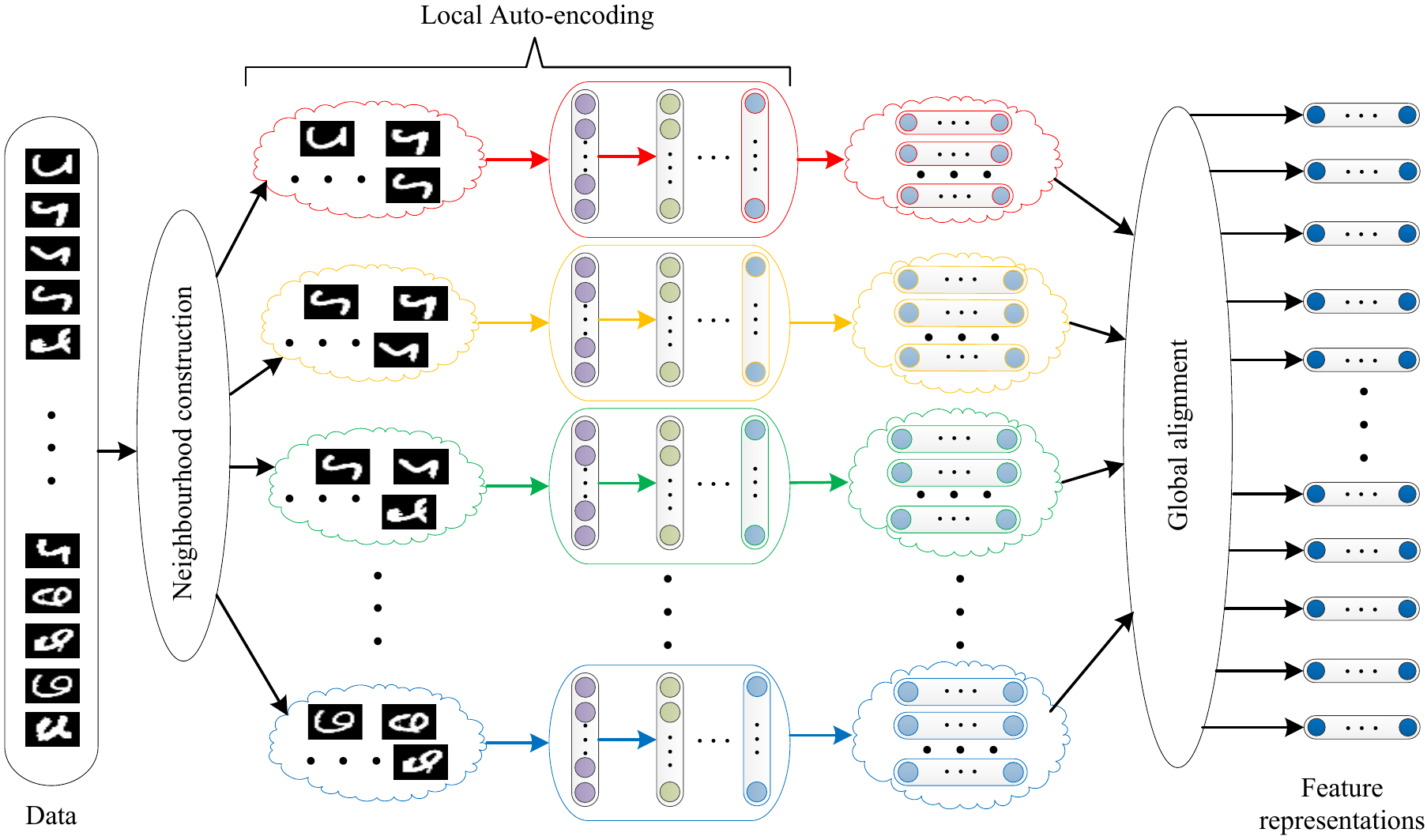}
\caption{ The basic framework of the proposed method: First, compute the neighbourhood for each data sample;
Second, extract deep-level features from each neighbourhood using the local SCAEs;
At last, align the local features to obtain the global feature representations.}
\label{fig:framework}
\end{figure*}


CAEs can be used as components to form a deep neural network, where
the lower layer's output serves as the higher and adjacent layer's
input. An example of a two-layer Stacked CAE (SCAE) is shown in
Fig. \ref{fig:sae} (b). We believe the SCAE will extract more
robust features than the one-layer CAE. The SCAE can be described
as:
\begin{align}\label{mxobjscae}
\begin{split}
& \min_{\bm{{\rm W}},\bm{{\rm b}},\bm{{\rm c}}} \sum_{l=1}^L \| \bm{{\rm H}}^l -g_{\bm{{\rm W}}^{(l)},\bm{{\rm b}}^{(l)}}(\bm{{\rm X}}^l)\|_F^2 +
\|\bm{{\rm X}}^l - g_{(\bm{{\rm W}}^{(l)})^T,\bm{{\rm c}}^{(l)}}(\bm{{\rm H}}^l)\|_F^2 \\
& + \lambda \trace \big ( (\bm{{\rm W}}^{(l)})^T (\bm{{\rm H}}^l \odot (\bm{{\rm E}} - \bm{{\rm H}}^l)) (\bm{{\rm H}}^l \odot
(\bm{{\rm E}} - \bm{{\rm H}}^l))^T \bm{{\rm W}}^{(l)}) \big ), \\
& s.t. \quad \bm{{\rm H}}^l = \bm{{\rm X}}^{l+1}, l = 1,\ldots,L-1,
\end{split}
\end{align}
where $\bm{{\rm X}}^l$ and $\bm{{\rm H}}^l$ represents the input and
output of the $l_{th}$ layer respectively, and the
superscript $(l)$ indicates that the parameters correspond to the
$l_{th}$ layer, and we reuse $\bm{{\rm W}}$, $\bm{{\rm b}}$ and
$\bm{{\rm c}}$ to represent the parameters of the SCAE.

\subsection{Objective Function of Local Deep-Feature Alignment}

Our basic concern is to extract deep-level features from each local
data subset that reflect some local characteristic of the data
subset. Then we align these local features to form
the global deep features.

To this end, we propose to construct a neighbourhood for each data
sample that includes the sample and a number of its closest
neighbouring samples. Then we use an SCAE to extract deep-level
features from each neighbourhood and impose a local affine
transformation on the deep features of each neighbourhood to align
the features from each local coordinate system with a global
coordinate system. The framework of the method is illustrated in
Fig. \ref{fig:framework}.

The objective function of the method should include two parts, local
deep-feature learning and global alignment of local features. For
each $\bm{{\rm x}}_i$, we define its neighbourhood $\bm{{\rm X}}_i$ as $\bm{{\rm X}}_i =
[\bm{{\rm x}}_i,\bm{{\rm x}}_{i_1},\ldots,\bm{{\rm x}}_{i_{k_i}}]$ where $k_i$ is its number of
neighbours. Hence the error of the SCAEs used for local feature
extraction from all the neighbourhoods can be represented as:
\begin{align}\label{localscae}
\begin{split}
& \sum_{i=1}^N \sum_{l=1}^L  \| \bm{{\rm H}}_{i}^l
-g_{\bm{{\rm W}}_{i}^{(l)},\bm{{\rm b}}_{i}^{(l)}}(\bm{{\rm X}}_{i}^l)\|_F^2 +
\|\bm{{\rm X}}_{i}^l - g_{(\bm{{\rm W}}_{i}^{(l)})^T,\bm{{\rm c}}_{i}^{(l)}}(\bm{{\rm H}}_{i}^l)\|_F^2 \\
& + \lambda \trace \big ( (\bm{{\rm W}}_{i}^{(l)})^T (\bm{{\rm H}}_{i}^l \odot (\bm{{\rm E}} -
\bm{{\rm H}}_{i}^l)) (\bm{{\rm H}}_{i}^l \odot
(\bm{{\rm E}} - \bm{{\rm H}}_{i}^l))^T \bm{{\rm W}}_{i}^{(l)}) \big ), \\
& s.t. \quad \bm{{\rm H}}_{i}^l = \bm{{\rm X}}_{i}^{l+1}, l = 1,\ldots,L-1
\end{split}
\end{align}
where the subscript $i$ is the index of the neighbourhood and other
symbols have the same meanings as in formula (\ref{mxobjscae}).

The top-layer local deep features $\bm{{\rm
H}}_{i}^L$ learned so far are neighbour-wise, and we need to derive
the global deep features. Based on the success of LTSA, it is
reasonable to assume that there exists an affine transformation
between $\bm{{\rm H}}_{i}^L$ and their global counterparts. Let
$\bm{{\rm A}}_i$ be the affine transformation matrix, the alignment
error of each neighbourhood can be described as
\begin{align}\label{erroralign}
  \|\bm{{\rm H}}_{i} \bm{{\rm T}}_i -  \bm{{\rm A}}_i \bm{{\rm H}}_{i}^L\|_F^2,
\end{align}
where $\bm{{\rm T}}_i = \bm{{\rm I}}_{k_i+1} -
\bm{{\rm e}}_{k_i+1} \bm{{\rm e}}_{k_i+1}^T/(k_i+1)$ moves the
feature representations in $\bm{{\rm H}}_{i}$ to their geometric
centre, $\bm{{\rm H}}_{i}$ is the global deep features corresponding
to the $i_{th}$ neighbourhood. We need to find
$\bm{{\rm A}}_i$ and $\bm{{\rm H}}_{i}$ such that $\bm{{\rm H}}_{i}$
preserves as much of the locality characteristics contained in
$\bm{{\rm H}}_{i}^L$ as possible. This problem can be solved by
minimizing the overall alignment error:
\begin{align}\label{objalign}
  \sum_{i=1}^N \|\bm{{\rm H}}_{i} \bm{{\rm T}}_i -  \bm{{\rm A}}_i \bm{{\rm H}}_{i}^L\|_F^2.
\end{align}

Combining (\ref{localscae}) and (\ref{objalign}) together, we
describe the proposed method as the following optimization problem:
\begin{align}\label{obj}
\begin{split}
& \min_{\bm{{\rm H}}_{i},\bm{{\rm W}}_{i},\bm{{\rm b}}_{i},\bm{{\rm
c}}_{i}}  \big ( \sum_{i=1}^N \sum_{l=1}^L
\| \bm{{\rm H}}_{i}^l -g_{\bm{{\rm W}}_{i}^{(l)},\bm{{\rm b}}_{i}^{(l)}}(\bm{{\rm X}}_{i}^l)\|_F^2 + \\
& \|\bm{{\rm X}}_{i}^l - g_{(\bm{{\rm W}}_{i}^{(l)})^T,\bm{{\rm c}}_{i}^{(l)}}(\bm{{\rm H}}_{i}^l)\|_F^2 + \\
& \lambda \trace \big ( (\bm{{\rm W}}_{i}^{(l)})^T (\bm{{\rm H}}_{i}^l \odot (\bm{{\rm E}} -
\bm{{\rm H}}_{i}^l)) (\bm{{\rm H}}_{i}^l \odot
(\bm{{\rm E}} - \bm{{\rm H}}_{i}^l))^T \bm{{\rm W}}_{i}^{(l)}) \big ) \big ) + \\
& \sum_{i=1}^N \|\bm{{\rm H}}_{i} \bm{{\rm T}}_i -  \bm{{\rm A}}_i \bm{{\rm H}}_{i}^L\|_F^2, \\
& s.t. \quad \bm{{\rm H}}_{i}^l = \bm{{\rm X}}_{i}^{l+1}, l = 1,\ldots,L-1
\end{split}
\end{align}
where $\bm{{\rm T}}_i = \bm{{\rm I}}_{k_i+1} - \bm{{\rm e}}_{k_i+1} \bm{{\rm e}}_{k_i+1}^T/(k_i+1)$.

\subsection{Optimization}

We adopt a two-stage strategy to optimize the problem (\ref{obj}).
In the first stage, we learn the local deep features using a series
of SCAEs. In the second stage, we align the local features to form
the global feature representations.

\begin{figure}[ht]
\centering
\includegraphics[width = 3in]{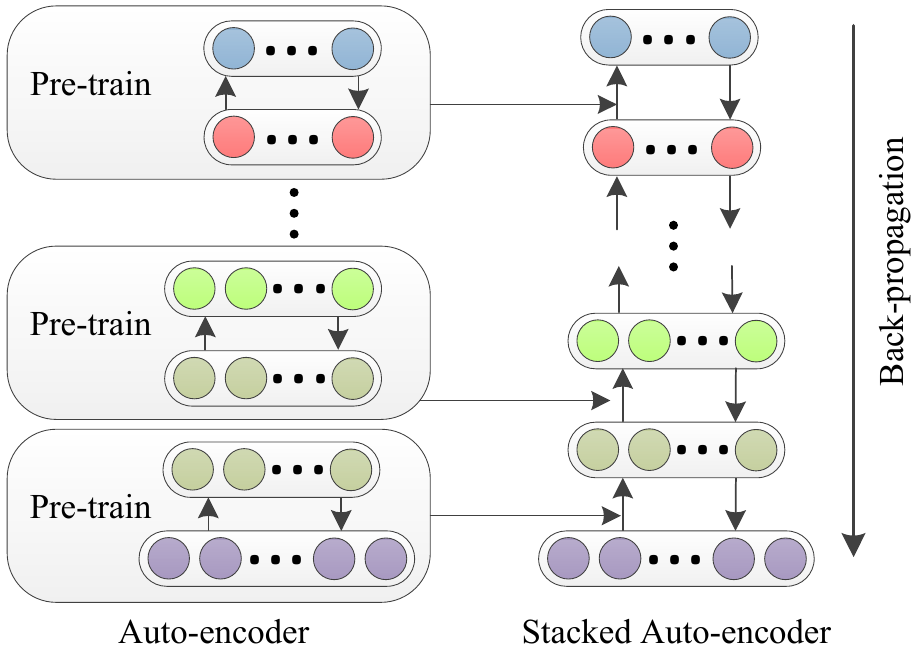}
\caption{ The optimization process of an SCAE: Pre-train each CAE layer by layer to obtain the initial parameters,
then initialize the SCAE using these parameters and fine-tune the SCAE using gradient descent algorithm with back-propagation.} \label{fig:optscae}
\end{figure}

\textbf{Stage 1:} In training each SCAE, to achieve optimized
encoding and decoding, we can separately pre-train each CAE and
optimize the deep network with the parameters initialized by the
pre-trained parameters of each layer. The problem can be solved
using a gradient descent algorithm with back-propagation
\cite{Chauvin1995Backpropagation}. The optimization process of an SCAE is shown in
Fig. \ref{fig:optscae}.

\textbf{Stage 2:} According to (\ref{objalign}), the
optimal-alignment matrix $\bm{{\rm A}}_i$ is given by $\bm{{\rm
A}}_i= \bm{{\rm H}}_{i} \bm{{\rm T}}_i (\bm{{\rm H}}_i^L)^\dag$
where $(\bm{{\rm H}}_i^L)^\dag$ is the Moore-Penrose generalized
inverse of $\bm{{\rm H}}_i^L$. Hence (\ref{objalign}) can be
rewritten as
\begin{align}\label{eqn:ltsaalign}
 \sum_{i=1}^N \|\bm{{\rm H}}_{i} \bm{{\rm T}}_i (\bm{{\rm I}}_{k_i+1} -
(\bm{{\rm H}}_i^L)^\dag \bm{{\rm H}}_i^L)\|_F^2.
\end{align}

Suppose $\bm{{\rm H}} = [\bm{{\rm h}}_1,\bm{{\rm
h}}_2,\dots,\bm{{\rm h}}_N] \in R^d$ are the final feature
representations where $\bm{{\rm h}}_i$ corresponds to $\bm{{\rm
x}}_i$, and let $\bm{{\rm S}}_i$ be the 0-1 selection matrix such
that $\bm{{\rm H}} \bm{{\rm S}}_i = \bm{{\rm H}}_{i}$. We then need
to find $\bm{{\rm H}}$ to minimize the overall alignment error:
\begin{align}\label{eqn:ltsaalign2}
  \sum_{i=1}^N \|\bm{{\rm H}}_{i} \bm{{\rm T}}_i (\bm{{\rm I}}_{k_i+1} -
(\bm{{\rm H}}_i^L)^\dag \bm{{\rm H}}_i^L)\|_F^2 = \|\bm{{\rm H}}\bm{{\rm S}}\bm{{\rm M}}\|_F^2,
\end{align}
where $\bm{{\rm S}} = [\bm{{\rm S}}_1,\ldots,\bm{{\rm S}}_N]$ and $\bm{{\rm M}} = \diag(\bm{{\rm M}}_1,\ldots,\bm{{\rm M}}_N)$ with
\begin{align}\label{eqn:M_i}
\bm{{\rm M}}_i = (\bm{{\rm I}}_{k_i+1} - \bm{{\rm e}}_{k_i+1}
\bm{{\rm e}}_{k_i+1}^T/(k_i+1))(\bm{{\rm I}}_{k_i+1}-(\bm{{\rm H}}_i^L)^\dag \bm{{\rm H}}_i^L).
\end{align}

Let $\bm{{\rm \Phi}} = \bm{{\rm S}}\bm{{\rm M}}\bm{{\rm M}}^T\bm{{\rm S}}^T$. Then, (\ref{eqn:ltsaalign2}) can be rewritten
as $trace(\bm{{\rm H}} \bm{{\rm \Phi}} \bm{{\rm H}}^T)$, thus we reformulate the problem as:
\begin{align}\label{eqn:eigen}
 \min_{\bm{{\rm H}}} trace(\bm{{\rm H}} \bm{{\rm \Phi}} \bm{{\rm H}}^T),
\end{align}
which is a typical eigenvalue problem and can be easily solved by
existing methods.

Given the high-dimensional data samples $\bm{{\rm X}} = [\bm{{\rm x}}_1,\bm{{\rm x}}_2,\dots,\bm{{\rm x}}_N]$,
the algorithm of the Local Deep Feature Alignment (LDFA) method can
be summarized as Algorithm \ref{algorithm1} to obtain the
low-dimensional feature representations $\bm{{\rm H}} =[\bm{{\rm h}}_1,\bm{{\rm h}}_2,\dots,\bm{{\rm h}}_N]$.

\begin{algorithm}
\caption{The LDFA algorithm for dimension reduction.}
\label{algorithm1}

Step 1: Compute the neighbourhood $\bm{{\rm X}}_i$ for each data sample $\bm{{\rm x}}_i$
based on the Euclidean distance metric;

Step 2: For each $\bm{{\rm X}}_i$, train a local deep SCAE: separately
pre-train each CAE, initialize the deep SCAE using the pre-trained
parameters of each layer and optimize the SCAE via the
gradient-descent algorithm with back-propagation;

Step 3: Compute the top-layer local deep features $\bm{{\rm H}}_i^L$ using each
SCAE's embedding process, which can be found in (\ref{localscae});

Step 4: Define $\bm{{\rm S}}_i$ via the index set of the data contained in
$\bm{{\rm X}}_i$, derive $\bm{{\rm M}}_i$ from $\bm{{\rm H}}_i^L$ following (\ref{eqn:M_i}), and
construct $\bm{{\rm S}}$ and $\bm{{\rm M}}$ ;

Step 5: Construct $\bm{{\rm \Phi}}$ using $\bm{{\rm S}}$ and $\bm{{\rm M}}$, and obtain $\bm{{\rm H}}$ by
solving the eigenvalue problem (\ref{eqn:eigen}).

\end{algorithm}

\subsection{Embedding a New Data Sample}
\label{embednewdata}

\begin{figure}
\centering
\includegraphics[width =1.5in]{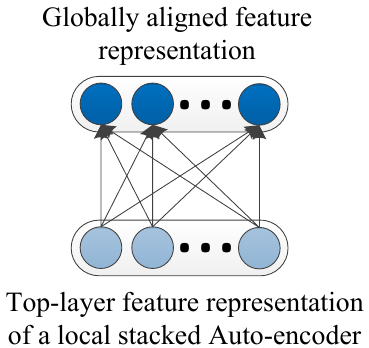}
\caption{ The fully connected network between the local feature
representation of a data sample and its global counterpart.}
\label{fig:embedalign}
\end{figure}

Our proposed method can be easily extended to embed a new data
sample into the learned low-dimensional subspace. We seek to
construct an explicit embedding function for each local
neighbourhood. Then, given a new data sample, we can find the
closest sample to it in the training set, and use the corresponding
embedding function to obtain the low-dimensional representation of
the new data sample.

To this end, we use a one-layer fully connected feed-forward neural
network to model the mapping from the top-layer local features
$\bm{{\rm H}}_i^L$ to the global feature representations $\bm{{\rm
H}}_{i}$. This one-layer network still exploits the sigmoid
activation function and its optimization can be described as:
\begin{align}\label{embedalign}
  \min_{\bm{{\rm \Theta}}_{i},\bm{{\rm u}}_{i}} \left\Vert \bm{{\rm H}}_{i} -  \frac{1}{1 +
  exp(-(\bm{{\rm \Theta}}_{i} \bm{{\rm H}}_i^L + \bm{{\rm u}}_i \bm{{\rm e}}^T))}  \right\Vert_{F}^2.
\end{align}
Fig. \ref{fig:embedalign} illustrates the one-layer fully
connected network.

\begin{figure}[h]
\centering
\includegraphics[width = 3in]{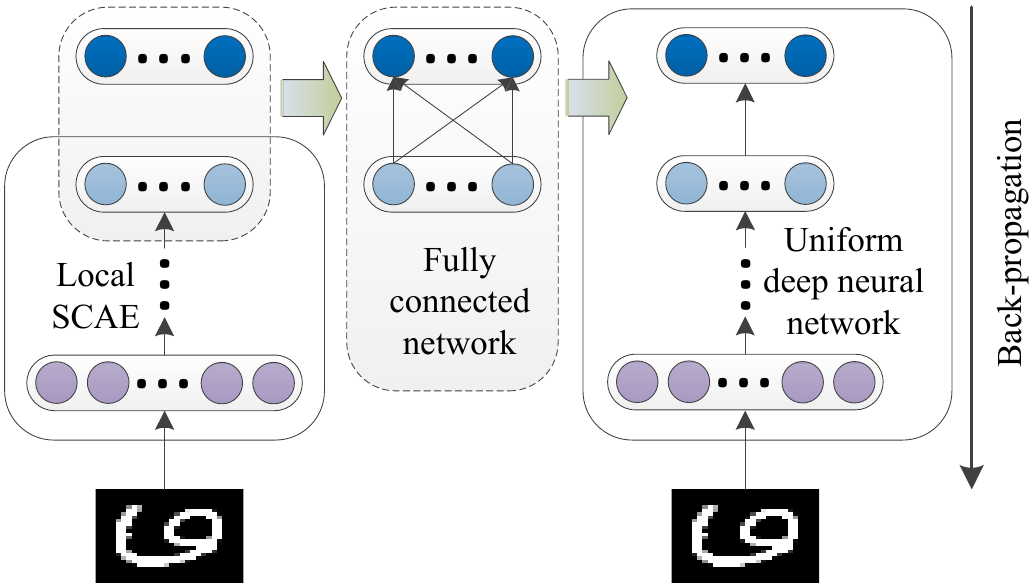}
\caption{ Construction of a new uniform deep neural network for
explicitly embedding data into a low-dimensional subspace.}
\label{fig:embeding}
\end{figure}

Next, we replace the local affine-transformation matrix between
$\bm{{\rm H}}_i^L$ and $\bm{{\rm H}}_{i}$ with the aforementioned
fully connected network so that it is stacked on
the top of the local SCAE, whose top-layer output
are $\bm{{\rm H}}_i^L$. This is shown in Fig. \ref{fig:embeding}.

Note that a CAE is also realized by a one-layer neural network,
which means that the fully connected network defined in
(\ref{embedalign}) shares the same mathematical form with a CAE. For
this reason, each SCAE, together with the corresponding fully
connected network, forms a new uniform deep neural network that is
able to explicitly embed a data sample into the learned
low-dimensional subspace to obtain the globally aligned feature
representation. The advantage of using the fully
connected network for local feature alignment is that we can
initialize the uniform deep neural network using the parameters of
the learned SCAE and fine-tune the uniform deep network via the
gradient-descent algorithm with back-propagation. The construction
of a uniform deep neural network is also shown in Fig.
\ref{fig:embeding}. It is worthwhile to point out that the
back-propagation can be used only when the local feature alignment
is achieved by fully connected neural network. Therefore Algorithm
\ref{algorithm1} does not include back-propagation in local feature
alignment.

Suppose each SCAE has $L$ layers. We build an $L+1$-layer deep
neural network for each neighbourhood, and then initialize the
parameters of the first $L$ layers with the trained SCAE and
initialize the parameters of the ${L+1}_{th}$ layer with the fully
connected network. Specifically, let $\bm{{\rm U}}^i(\cdot)$ be the
explicit embedding function representing the uniform deep neural
network, such that $\bm{{\rm U}}_{\bm{{\rm Q}}_i,\bm{{\rm
v}}_i}^i(\bm{{\rm X}}_i)=\bm{{\rm H}}_{i}$. We initialize its
parameters $\bm{{\rm Q}}_i$ and $\bm{{\rm v}}_i$ in the following
way.
\begin{eqnarray}\label{initialization}
\left \{
\begin{aligned}
  & {\bm{{\rm Q}}_i}^{(l)} = {\bm{{\rm W}}_i}^{(l)},\quad {\bm{{\rm v}}_i}^{(l)} = {\bm{{\rm b}}_i}^{(l)}, \quad (l=1,\ldots,L) \\
  & {\bm{{\rm Q}}_i}^{(l)} = \bm{{\rm \Theta}}_{i},\quad {\bm{{\rm v}}_i}^{(l)} = \bm{{\rm u}}_i, \quad l = L+1, \\
\end{aligned}
\right.
\end{eqnarray}
where the superscript $(l)$ indicates the parameters in the $l_{th}$
layer of the uniform deep network. Once $\bm{{\rm U}}^i(\cdot)$ is obtained, we
can locate the nearest neighbour of a new data sample in the
training set, and use the corresponding $\bm{{\rm U}}^i(\cdot)$ to embed the
data sample into a low-dimensional subspace.

To realize the embedding of a new data sample $\bm{{\rm \tilde x}}$, we modify the original LDFA algorithm such that it splits into Algorithm 2 and Algorithm 3, which describe the training and embedding process respectively.

\begin{algorithm}[htb]
\caption{Training of LDFA for out-of-sample data embedding.}
\label{algorithm2}

Step 1: Compute the neighbourhood $\bm{{\rm X}}_i$ for each data sample $\bm{{\rm x}}_i$
based on the Euclidean distance metric;

Step 2: For each $\bm{{\rm X}}_i$, train a local deep SCAE: separately
pre-train each CAE, initialize the deep SCAE using the pre-trained
parameters of each layer and optimize the SCAE via the
gradient-descent algorithm with back-propagation;

Step 3: Compute the top-layer local deep features $\bm{{\rm H}}_i^L$ using each
SCAE's embedding process, which can be found in (\ref{localscae});

Step 4: Define $\bm{{\rm S}}_i$ via the index set of the data contained in
$\bm{{\rm X}}_i$, derive $\bm{{\rm M}}_i$ from $\bm{{\rm H}}_i^L$ following (\ref{eqn:M_i}), and
construct $\bm{{\rm S}}$ and $\bm{{\rm M}}$ ;

Step 5: Construct $\bm{{\rm \Phi}}$ using $\bm{{\rm S}}$ and $\bm{{\rm M}}$, obtain $\bm{{\rm H}}$ by solving
the eigenvalue problem (\ref{eqn:eigen});

Step 6: Build a one-layer fully connected neural network between
each pair $\bm{{\rm H}}_i^L$ and $\bm{{\rm H}}_{i}$ according to
(\ref{embedalign});

Step 7: Initialize the uniform deep neural networks $\bm{{\rm U}}^i(\cdot)$
through (\ref{initialization}) and fine-tune the networks using
back-propagation.

\end{algorithm}

\begin{algorithm}[htb]
\caption{Out-of-sample embedding using trained LDFA.}
\label{algorithm3}

Step 1: Locate $\bm{{\rm \tilde x}}$'s nearest neighbour $\bm{{\rm x}}_j$ in $\bm{{\rm X}}$, and then
obtain the neighbourhood $\bm{{\rm X}}_j$;

Step 2: Use $\bm{{\rm U}}^j(\cdot)$ to embed $\bm{{\rm \tilde x}}$ into a low-dimensional
subspace and obtain its low-dimensional feature representation.

\end{algorithm}



\section{Experiments}
\label{exp}

We will use the proposed LDFA method in several representative
applications and evaluate its performances. Dimension reduction is
commonly used as a preprocessing step for subsequent
data-visualization, data-clustering and data-classification tasks.
Therefore, we will first examine LDFA's image-visualization
capability and clustering accuracy based on the images that
have been dimension reduced using LDFA. In
addition, we will determine the classification accuracy based on the
LDFA feature representations.
Both qualitative and quantitative
experimental results will be reported, and a comparison with other
existing methods will be provided.

The clustering accuracy is defined as the purity, which is computed
as the ratio between the number of correctly clustered samples and
the total number of samples:
\begin{align*}
purity(\Omega,C) = \frac{1}{N} \sum_{i} \max_{j}
\{\omega_{i},c_{j}\}, \quad \\ i = 1,\dots,N_c, j = 1, \dots,N_c,
\end{align*}
where $\Omega = \{\omega_{1},\ldots,\omega_{N_c}\}$ is the clustered
data set with $\omega_{i}$ representing the data in the $i_{th}$
cluster and $C = \{c_1,\ldots,c_{N_c}\}$ is the original data set
with $c_j$ representing the data in the $j_{th}$ class.

\subsection{The Data Sets}
\label{data}

The experiments adopt seven benchmark data sets for
image visualization/clustering/classification, the data sets include
the MNIST Digits \footnotemark[1]
\footnotetext[1]{\url{http://www.cs.nyu.edu/~roweis/data.html}},
USPS Digits \footnotemark[1], Olivetti Faces \footnotemark[1], the
UMist Faces \footnotemark[1], the NABirds
\footnotemark[2]\footnotetext[2]{http://dl.allaboutbirds.org/nabirds},
the Stanford Dogs
\footnotemark[3]\footnotetext[3]{http://vision.stanford.edu/aditya86/ImageNetDogs/},
and the Caltech-256
\footnotemark[4]\footnotetext[4]{http://www.vision.caltech.edu/ImageDatasets/Caltech256/intro/}
data sets.

Table \ref{tab:bench} shows the attributes of these
data sets and how we use these data sets in the experiments. The
attributes of each data set are the class number, the total number
of data and the data dimension. Considering the computational
efficiency, we are not going to use all the data for evaluation.
Table \ref{tab:bench} clearly indicates how many images (and per
class) are involved in the experiments, and how many images are
chosen for training and testing respectively. Particularly, the
NABirds data set covers 400 species, but only 100 species are
involved in our experiments. All the experiments are repeated 10
times, with randomly selected images in each time, and we show the
statistical results of these experiments using box plot.

\begin{figure*}
\centering
\includegraphics[width=7in, bb=80 2 984 457]{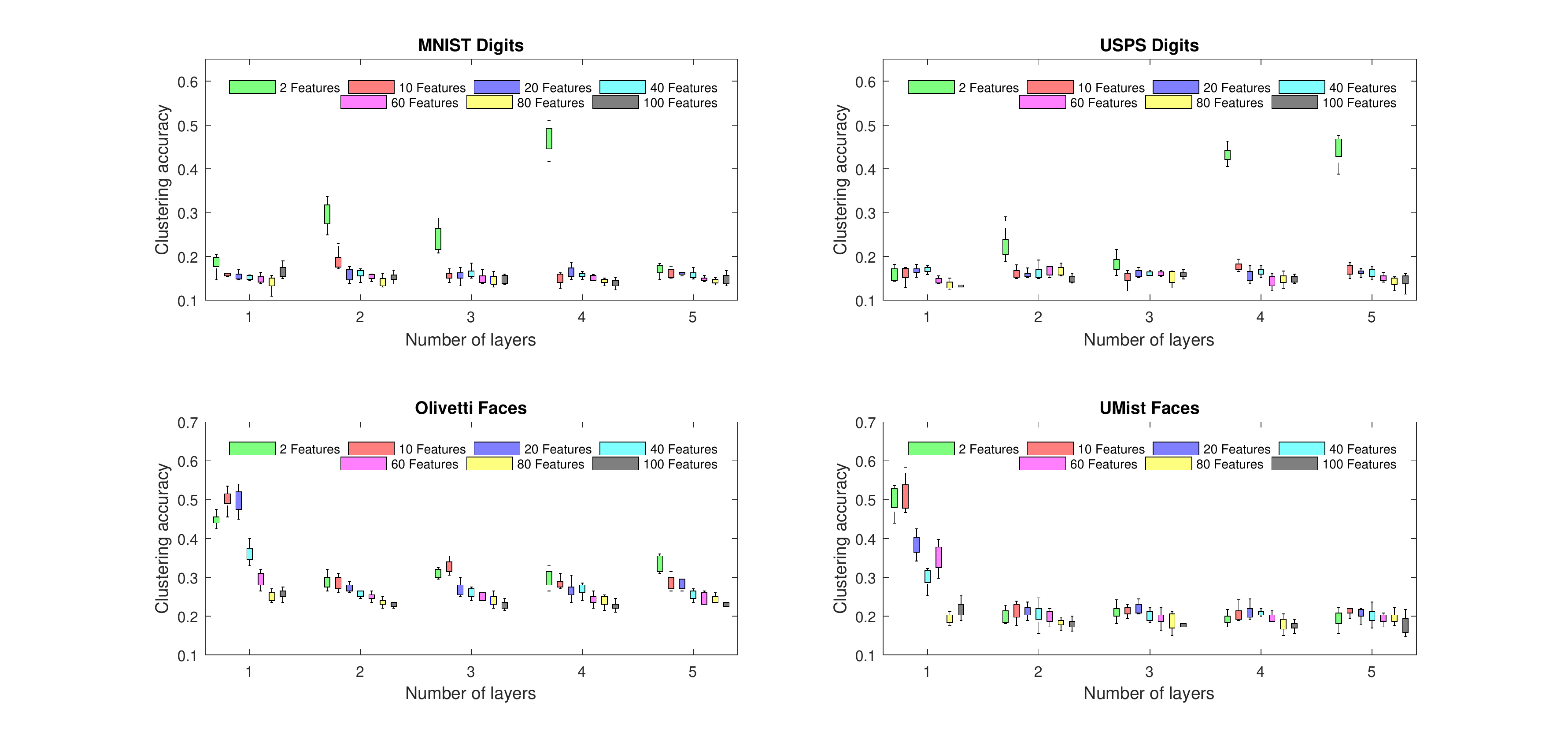}
\caption{ Clustering accuracy of the LDFA method with different numbers of layers of local SCAEs and different feature dimensions.}
\label{fig:layerdim}
\end{figure*}

\begin{table*}
\caption{A brief description of the data sets used
in the experiments.} \label{tab:bench}
\begin{center}
\newcommand{\tabincell}[2]{\begin{tabular}{@{}#1@{}}#2\end{tabular}}
  \centering
  \begin{tabular}{ l c c c c c c c}\hline
   Data Sets & \tabincell{c}{Class \\ number} & \tabincell{c}{Total number \\ of data} & \tabincell{c}{Data \\ dimension} & \tabincell{c}{Number of \\ data in use}  & \tabincell{c}{Number of \\ data per class} & \tabincell{c}{Number of training\\ data per class} & \tabincell{c}{Number of testing\\ data per class}  \\\hline
   MNIST Digits  &  10 & 70000 & 784 & 1000 & 100 & 70 & 30 \\
   USPS Digits &  10 & 11000 & 256 & 1000 & 100 & 70 & 30 \\
   Olivetti Faces & 40 & 400 & 4096 & 200 & 5 & 4 & 1 \\
   UMist Faces & 20 & 575 & 10304 & 360 & 18 & 15 & 3 \\
   NABirds & 400 & 48000 & variable & 3000 & 30 & 20 & 10  \\
   Stanford Dogs & 120 & 20580 & variable & 3600 & 30 & 20 & 10 \\
   Caltech-256 & 257 & 30608 & variable & 1,2850 & 50 & 30 & 20 \\
  \hline
\end{tabular}
\end{center}
\end{table*}

\subsection{Data Visualization and Clustering}
\label{visualization}

In data visualization, the original data are embedded into a two- or
three-dimensional subspace and the low-dimensional embeddings are
rendered to show the spatial relationships between data samples. A
good visualization result usually groups data of the same class
together and separates data from different classes. In this sense, a
good data-visualization result leads to high data-clustering
accuracy, and viceversa. The result of data visualization and the
accuracy of clustering reflect the discriminative information
contained in the data, so they are commonly used for evaluating the
feature representations learned by unsupervised
dimension reduction algorithms.

Three factors may influence the clustering performance of the LDFA
algorithm: the number of layers of the local SCAEs, the dimension of
the output feature representations, and the size of the
neighbourhood. Hence we want to find the proper number of layers,
feature dimensions and neighbourhood size for the LDFA algorithm.
First, we fix the neighbourhood size to 10, which often generates
good results in many locality-preserving dimension reduction
algorithms \cite{smith2008convergence}, and run Algorithm
\ref{algorithm1} dozens of times with different combinations of
numbers of layers and feature dimensions on the MNIST Digits data
set, USPS Digits data set, Olivetti Faces data set and UMist Faces
data set. We perform K-means clustering \cite{hartigan1979algorithm}
using the output feature representations and compute the clustering
accuracies. The results of ten repeated experiments are shown in Fig. \ref{fig:layerdim}, 
where we find that four- to five-layer local SCAEs with
two-dimensional features are sufficient to obtain good result for
digit data sets, while one-layer local SCAEs with two- to
ten-dimensional features are sufficient to get good result for face
data set. Specifically, the appropriate network structures of the
local SCAEs applied to the MNIST Digits, USPS Digits, UMist Faces
and Olivetti Faces data sets for data visualization can be
784-300-200-150-2, 256-300-250-200-150-2, 10304-2, and 4096-2,
respectively. The number of neurons in the network are smaller than
that used in \cite{hinton2006reducing}; we believe the reason for
this is that LDFA learns features from neighbourhoods, which usually
contain only very similar data samples, thus it does not need a very
complex network structure.

\begin{figure}
\centering
\includegraphics[width = 3.2in]{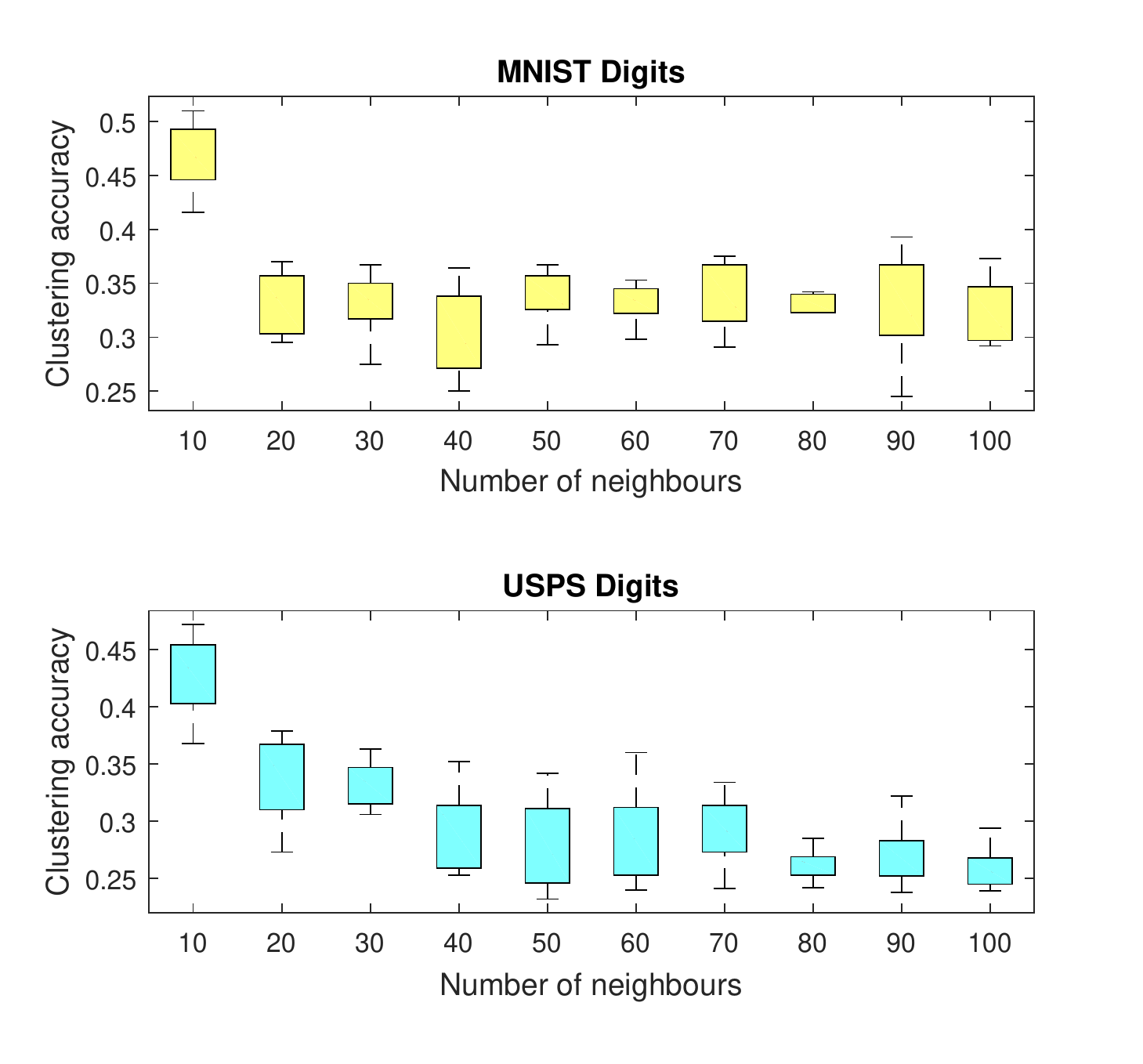}
\caption{Clustering accuracy of the LDFA method on
the MNIST data set and USPS data set with neighbourhood sizes
ranging from 10 through 100.} \label{fig:accu_neighbor}
\end{figure}

Then, using the network depth described above, we change the
neighbourhood size in Algorithm \ref{algorithm1} from 10 to 100 by
intervals of 10, and the clustering accuracies of ten repeated
experiments on the MNIST and USPS data sets are shown in Fig.
\ref{fig:accu_neighbor}. We find that the performance of the LDFA
algorithm drops dramatically when the neighbourhood size increases
from 10 to 20. Therefore, we believe LDFA learns discriminative
local features well with relatively small neighbourhood size.

\begin{figure}
\centering
\includegraphics[width = 3.2in]{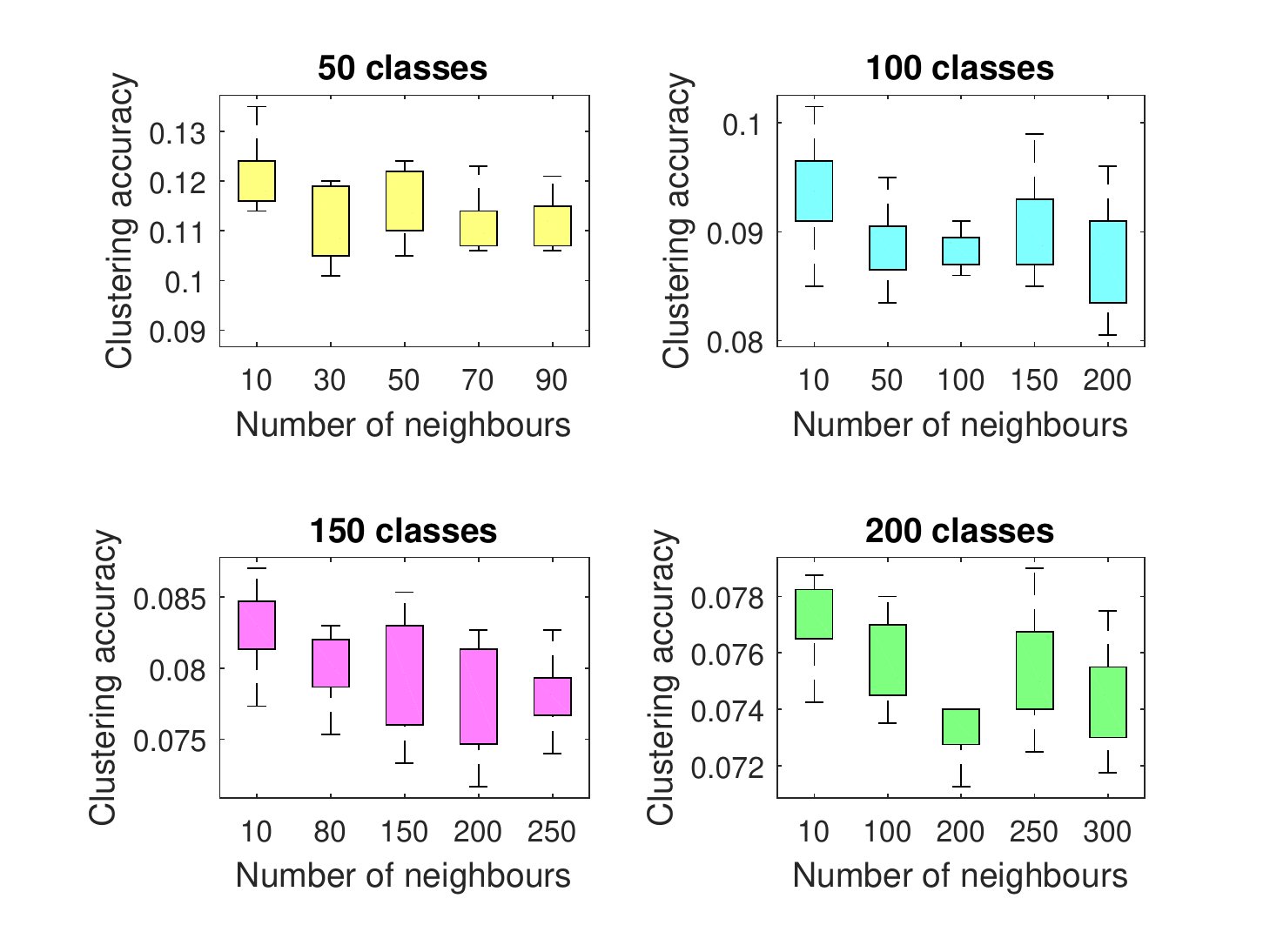}
\caption{Clustering accuracy of the LDFA method on
the Caltech-256 data set with different number of classes and
neighbourhood sizes.} \label{fig:accu_neighbor2}
\end{figure}

To further testify our speculation about the
neighbourhood size, we perform another four tests on Caltech-256
data set that covers much more classes than MNIST and USPS data
sets. In the first test, we choose 1000 samples from 50 classes with
20 samples per class and extract 59-dimension Local Binary Pattern
(LBP) features \cite{ahonen2006face} from these images. Then we
compute the clustering accuracies based on the two dimensional
features extracted by the LDFA, with the neighbourhood size varying
from 10 to 90. In the second test, 2000 samples are chosen from 100
classes, and the neighbourhood size varies from 10 to 200. In the
third test, 3000 samples are chosen from 150 classes, and the
neighbourhood size varies from 10 to 250. In the last test, 4000
samples are selected from 200 classes, and the neighbourhood size
varies from 10 to 300. The network structure is 59-30-10-2, and the
experiments are also repeated 10 times with randomly selected
samples each time. The clustering accuracies of these four tests are
shown in Fig. \ref{fig:accu_neighbor2} where we find the clustering
accuracies are not increasing with bigger neighbourhood sizes when
more classes are given.

\begin{figure}
\flushleft
\includegraphics[width = 3.5in]{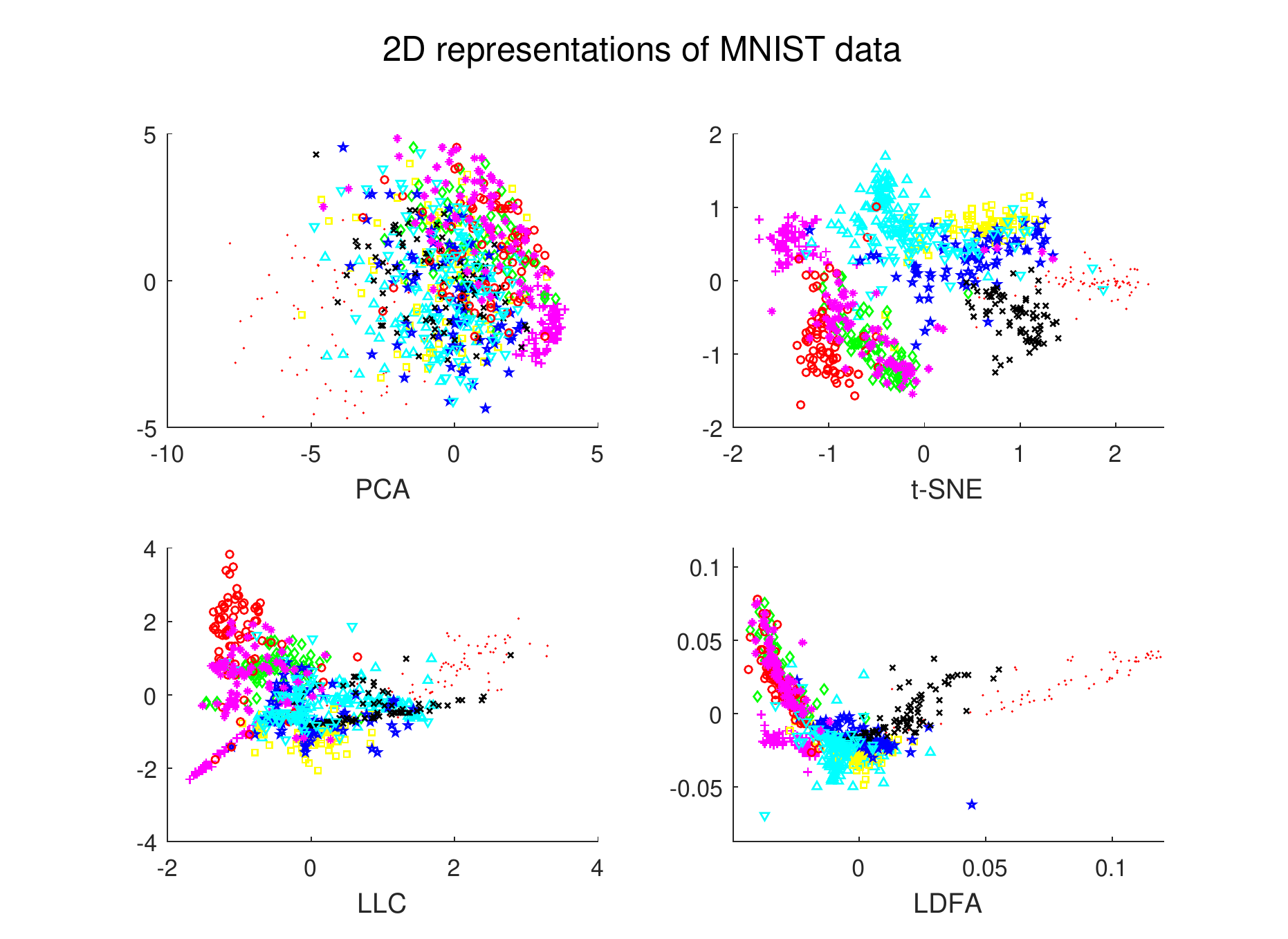}
\caption{The 2-D visualization results of the MNIST
Digits data set by PCA, t-SNE, LLC and LDFA.} \label{fig:vis_mnist}
\end{figure}

\begin{figure}[!htb]
\centering
\includegraphics[width = 3.5in]{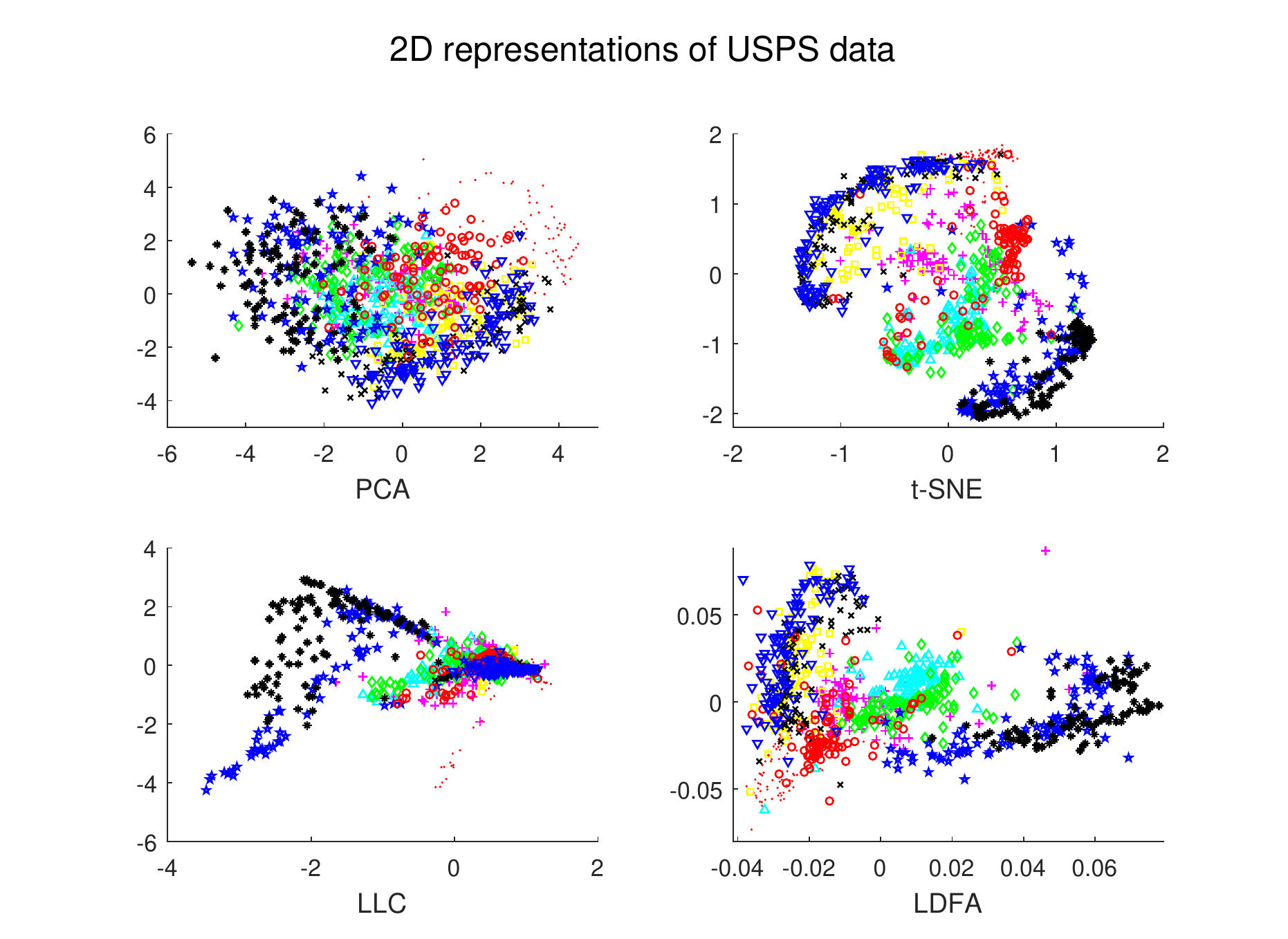}
\caption{The 2-D visualization result of the USPS
Digits data set by PCA, t-SNE, LLC and LDFA.} \label{fig:vis_usps}
\end{figure}

\begin{figure}
\centering
\includegraphics[width = 3.5in]{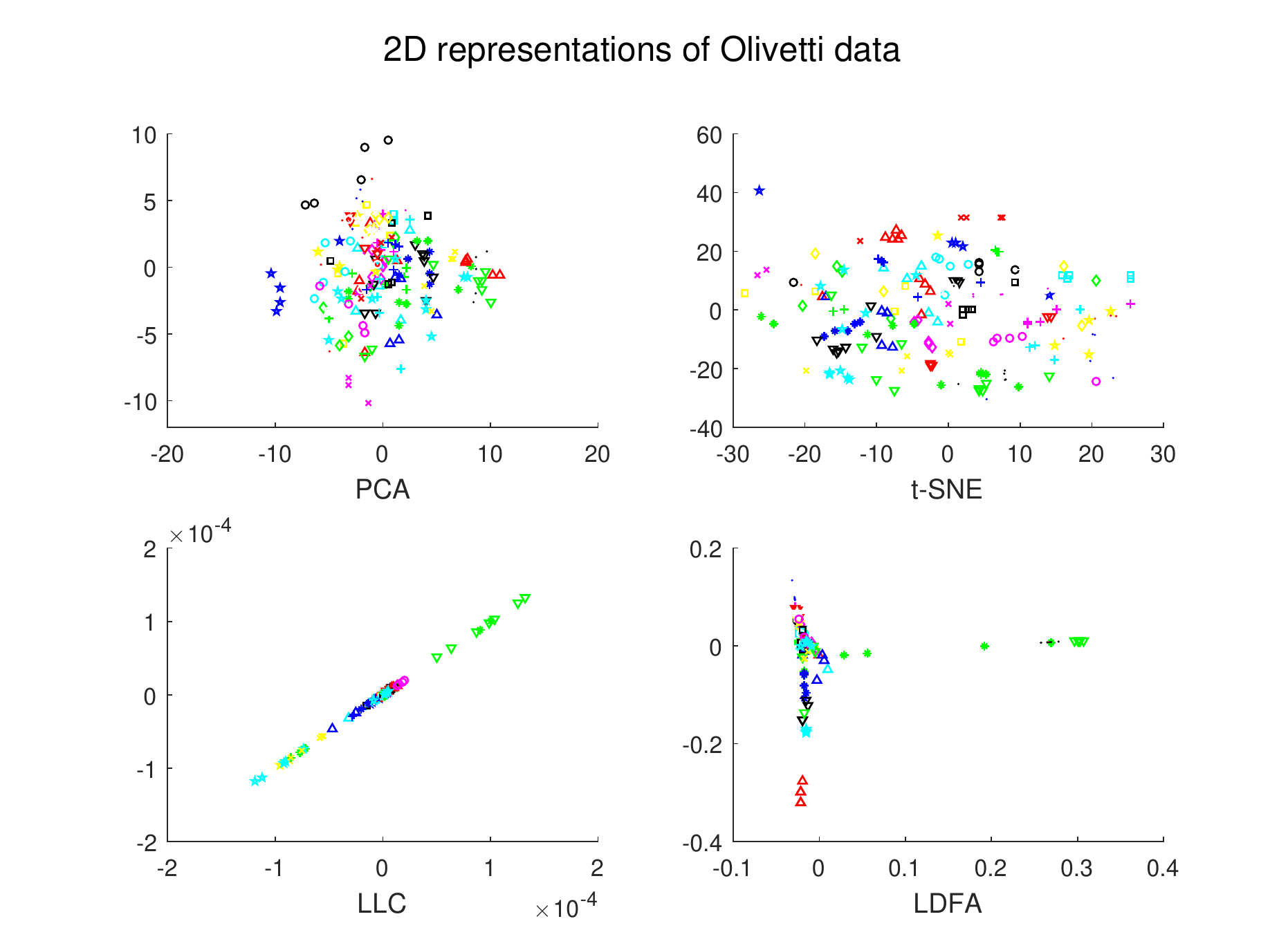}
\caption{The 2-D visualization result of the
Olivetti Faces data set by PCA, t-SNE, LLC and LDFA.}
\label{fig:vis_olive}
\end{figure}

\begin{figure}
\centering
\includegraphics[width = 3.5in]{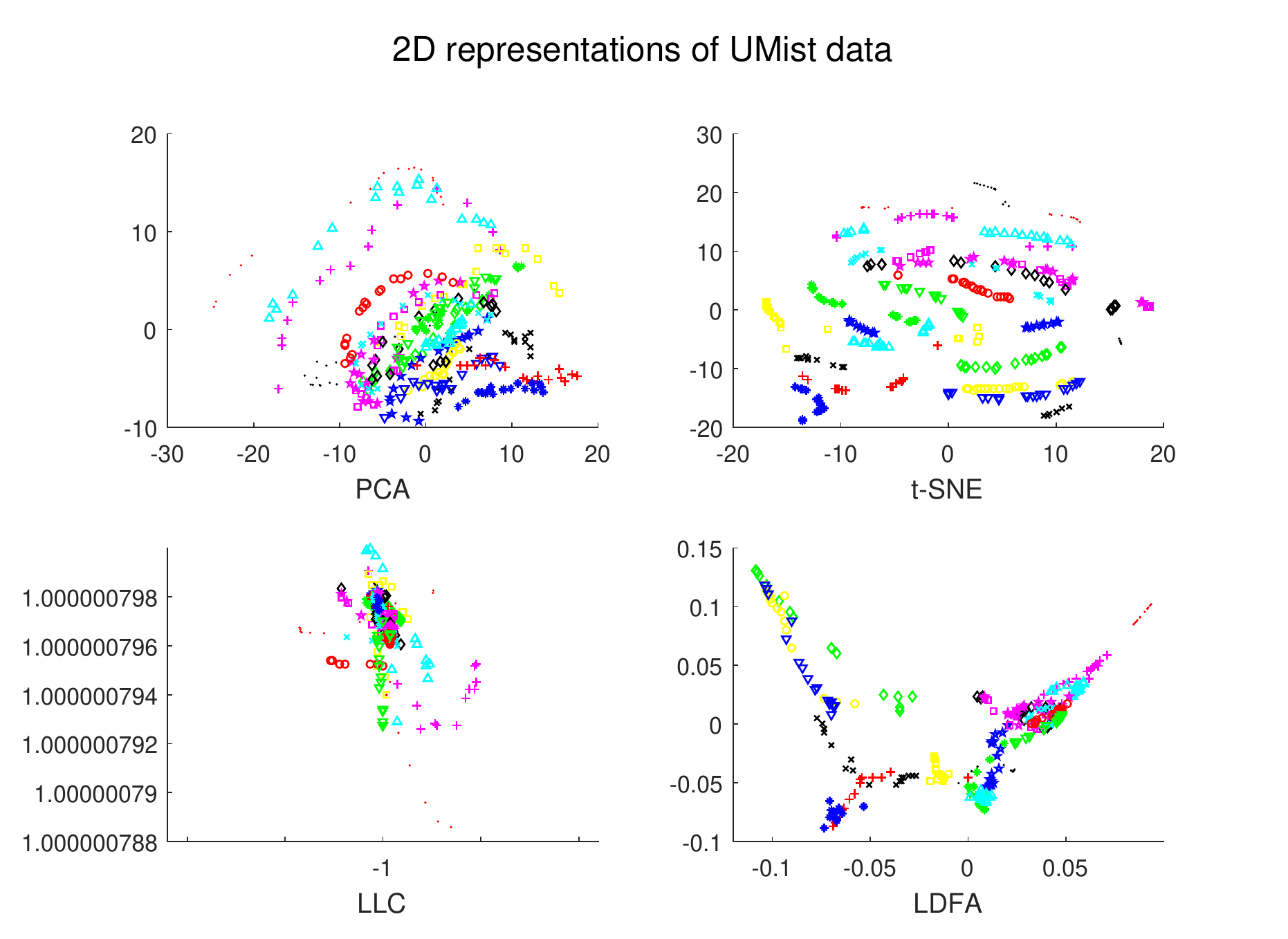}
\caption{The 2-D visualization result of the UMist
Faces data set by PCA, t-SNE, LLC and LDFA.} \label{fig:vis_umist}
\end{figure}

Consequently, we fix the neighbourhood size in the
LDFA algorithm to 10 to reduce the dimension of the aforementioned
four data sets. The 2-D visualization results of the four data sets
are depicted in Fig. \ref{fig:vis_mnist}, Fig. \ref{fig:vis_usps},
Fig. \ref{fig:vis_olive} and Fig. \ref{fig:vis_umist}. For
comparison, we also demonstrate the visualization results of three
other methods: the PCA, the t-SNE \cite{maaten2008visualizing} and
the Locally Linear Coordination (LLC)
\cite{roweis2002global}\footnotemark[5]\footnotetext[5]{http://lvdmaaten.github.io/drtoolbox/}.
The t-SNE solves a problem known as the crowding problem during
dimension reduction, which could be severe when the embedding
dimension is very low. Therefore it is anticipated that the 2-D
features learned by the t-SNE can well reflect the real data
distribution. Fig. \ref{fig:vis_mnist} through Fig.
\ref{fig:vis_umist} show that the LDFA are close to or comparable
with t-SNE in data visualization, but better than the PCA and the
LLC.
This is owing to the ability of the LDFA to capture not only the
global but also the local deep-level characteristics of the data
sets.



\begin{figure}
\centering
\includegraphics[trim = 19mm 0mm 0mm 0mm, clip=true, width = 3.9in]{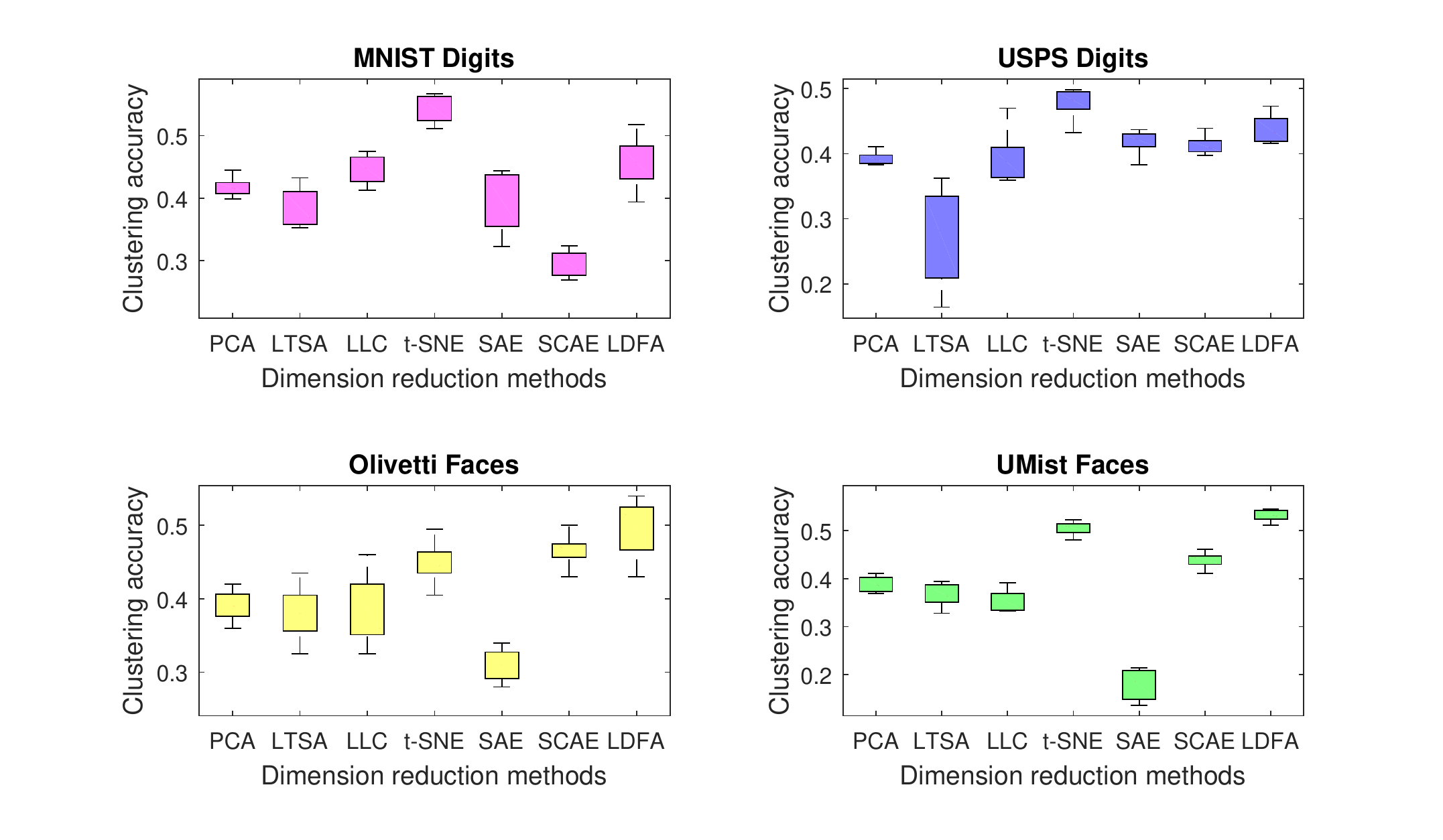}
\caption{The clustering accuracies base on the
dimension reduction results of PCA, LTSA, LLC, t-SNE, SAE, SCAE and
LDFA.} \label{fig:clu_four}
\end{figure}

To quantitatively evaluate the dimension reduction
results of Fig. \ref{fig:vis_mnist} through Fig.
\ref{fig:vis_umist}, we show the clustering accuracies based on
those 2-D features in Fig. \ref{fig:clu_four}, where we also show
the clustering accuracies derived from the LTSA, the basic stacked
Auto-encoders (SAE) and the SCAE. Different from other methods, the
features learned by the SAE and
SCAE for clustering are of 30 dimensions for good performance.
In this experiment, we apply the same network structure to the SAE,
SCAE and LDFA except for the dimension of the output data. The
clustering on each data set is also repeated 10 times. On MNIST and
USPS data sets, the LDFA is better than the PCA, LTSA, LLC, SAE and
SCAE owing to the ability to capture local deep features, but
inferior to the t-SNE, whose good performance is predictable from
the dimension reduction results shown in Fig. \ref{fig:vis_mnist}
through Fig. \ref{fig:vis_umist}. On Olivetti and UMist Faces data
sets, the LDFA achieves the best performances. It is noteworthy that
the SCAE performs worse than the SAE on MNIST and USPS data sets. We
think this can be explained by the characteristic of the SCAE's
regularization term to ignore the data variations that are
significant even for the same kind of digits. On Olivetti and UMist
Faces data sets, the performance of the SCAE is much better than the
SAE because human faces share very similar structure and
differenciate from one another mainly in texture, thus are more
suitable to be processed by the SCAE. Also, we believe the training
of the SAE needs relatively large sample set because the SAE tends
to fail in capturing the data variations when the sample set is
small. This explains why the clustering accuracies of the SAE on
Olivetti and UMist Faces data sets are low in Fig.
\ref{fig:clu_four}.

\begin{figure}
\centering
\includegraphics[width = 3.5in]{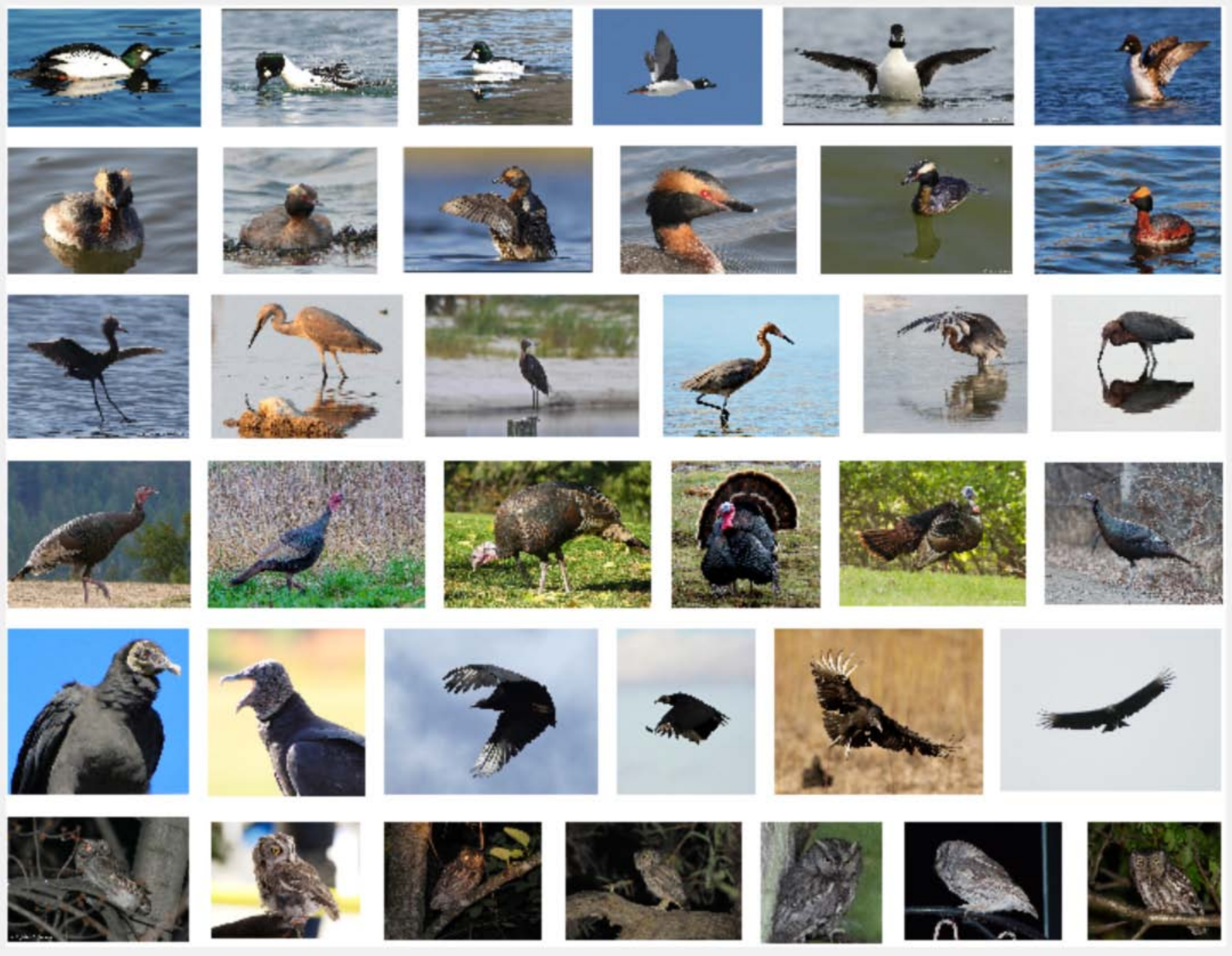}
\caption{ Some sample images selected from the NABirds data set.}
\label{fig:bird}
\end{figure}

\begin{figure}
\centering
\includegraphics[width = 3 in]{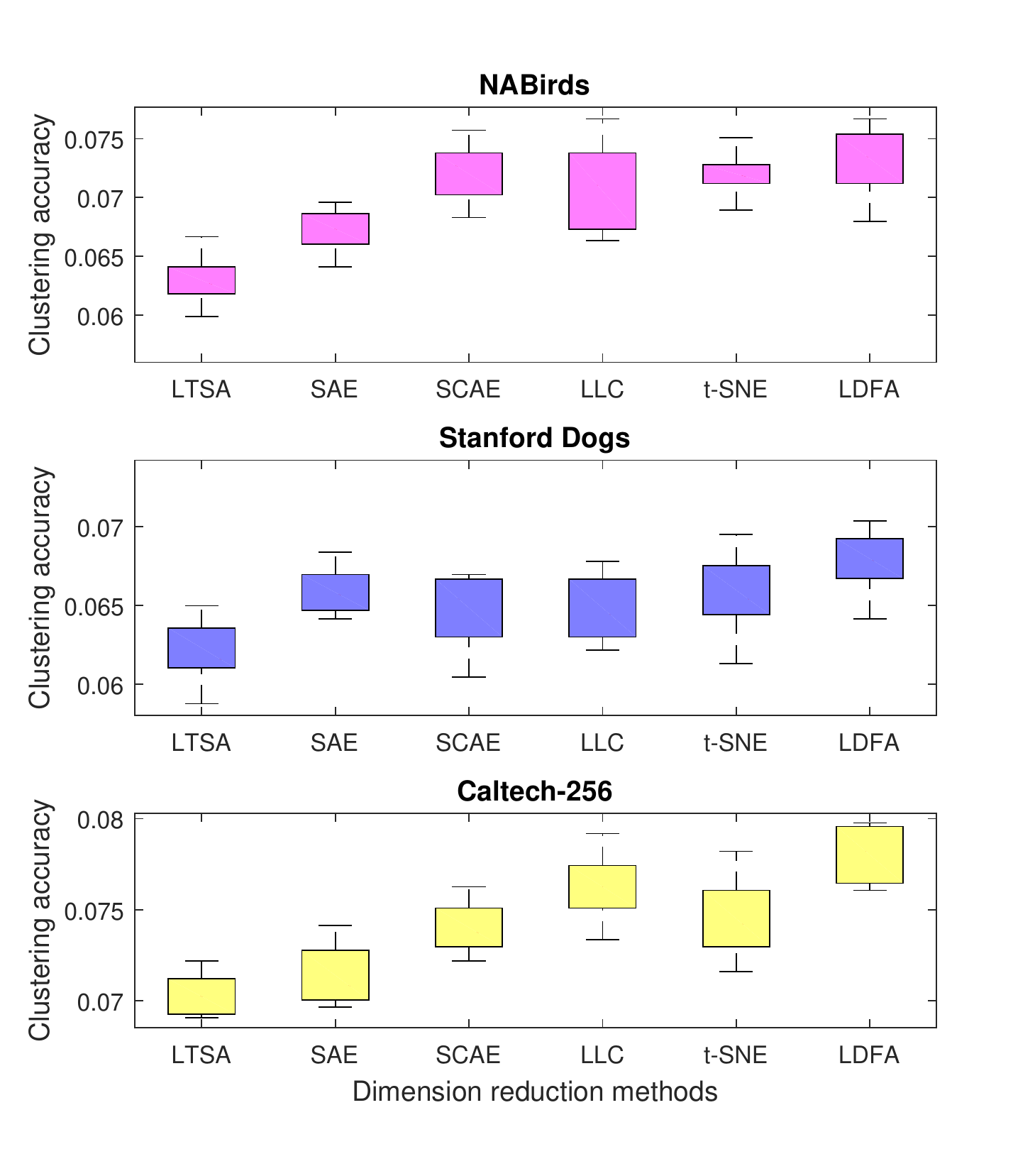}
\caption{Clustering accuracies of LTSA, SAE,
SCAE,LLC,t-SNE and LDFA on three bigger data sets.}
\label{fig:cluster_three}
\end{figure}

Additionally, we perform image clustering on three bigger data
sets-the NABirds, Stanford Dogs and Caltech-256, using the
low-dimensional features produced by the aforementioned dimension
reduction algorithms.
These three data sets are much more challenging for feature learning
because the foreground objects in the images are usually shown in
different poses and sizes, and a large number of images are
cluttered with natural scenes in the background. Some of the sample
images randomly selected from the NABirds data set are shown in
Fig.\ref{fig:bird}, where each row represents a breed of birds.
It is obvious that there exists great differences
between the images of the same class. In order to improve the
robustness, we extract a 59-dimension LBP feature descriptor from
each image to represent it. For the SAE, SCAE and LDFA, we use the
same network structure. For the LTSA, LLC and t-SNE, we use the
default settings in the existing implementations. We repeat the
experiment 10 times. The clustering accuracies derived from these
dimension reduction algorithms are depicted in Fig.
\ref{fig:cluster_three} where the LDFA outperforms the other five
algorithms.


\subsection{Data Classification}

To further evaluate the discriminative information contained in the
low-dimensional feature representations, we also perform image
classification using the dimension-reduced data. The classification
algorithms are carefully tuned so that they can generate their best
results. Note that we aim to compare the feature representations
learned by different methods, not to achieve the highest possible
classification accuracy. Thus, we deliberately do not use some
state-of-the-art classification methods.

\begin{figure*}
\centering
\includegraphics[width = 5.5in]{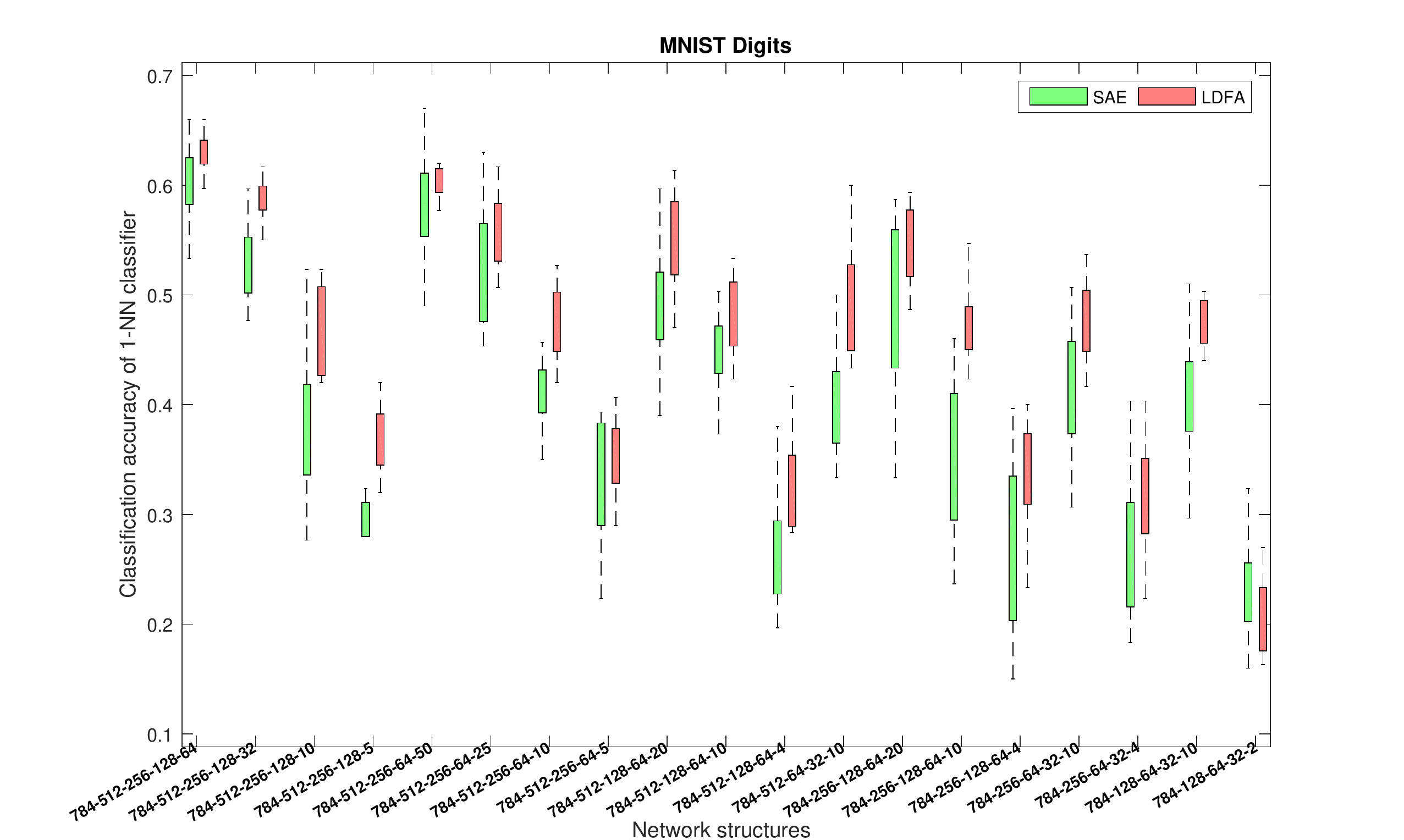}
\caption{Classification accuracies derived from SAE
and LDFA using different network structures on the MNIST data set.}
\label{fig:mniststructure}
\end{figure*}

Designating a very low data dimension in learning
might cause the features extracted by the AEs to "collapse" onto the
same dimension \cite{rifai2011contractive}, and this will
undoubtedly influence the classification accuracy. We believe
similar problem may happen to the SAE. Hence, we need to redesign
the network structure for data classification. As pointed out by
\cite{van2009dimensionality}, four layers should be appropriate for
an AE to learn good features. So we define 19 different four-layer
network structures and implement the SAE using these structures to
extract features, which are then fed into a 1-NN classifier. The
classification is performed 10 times and the results are show in
Fig.\ref{fig:mniststructure} where the SAE with a 784-512-256-128-64
structure generates the best results. In addition, we plot the
classification results derived from the LDFA using the 19
structures. It is clear that the LDFA outperforms the SAE with the
same network structure. We will not raise the dimension of the
feature representations any higher to prevent introducing the noise
and instability into the feature learning \cite{levina2005maximum}.

\begin{figure*}[]
\includegraphics[width=7in, bb=0 20 900 420]{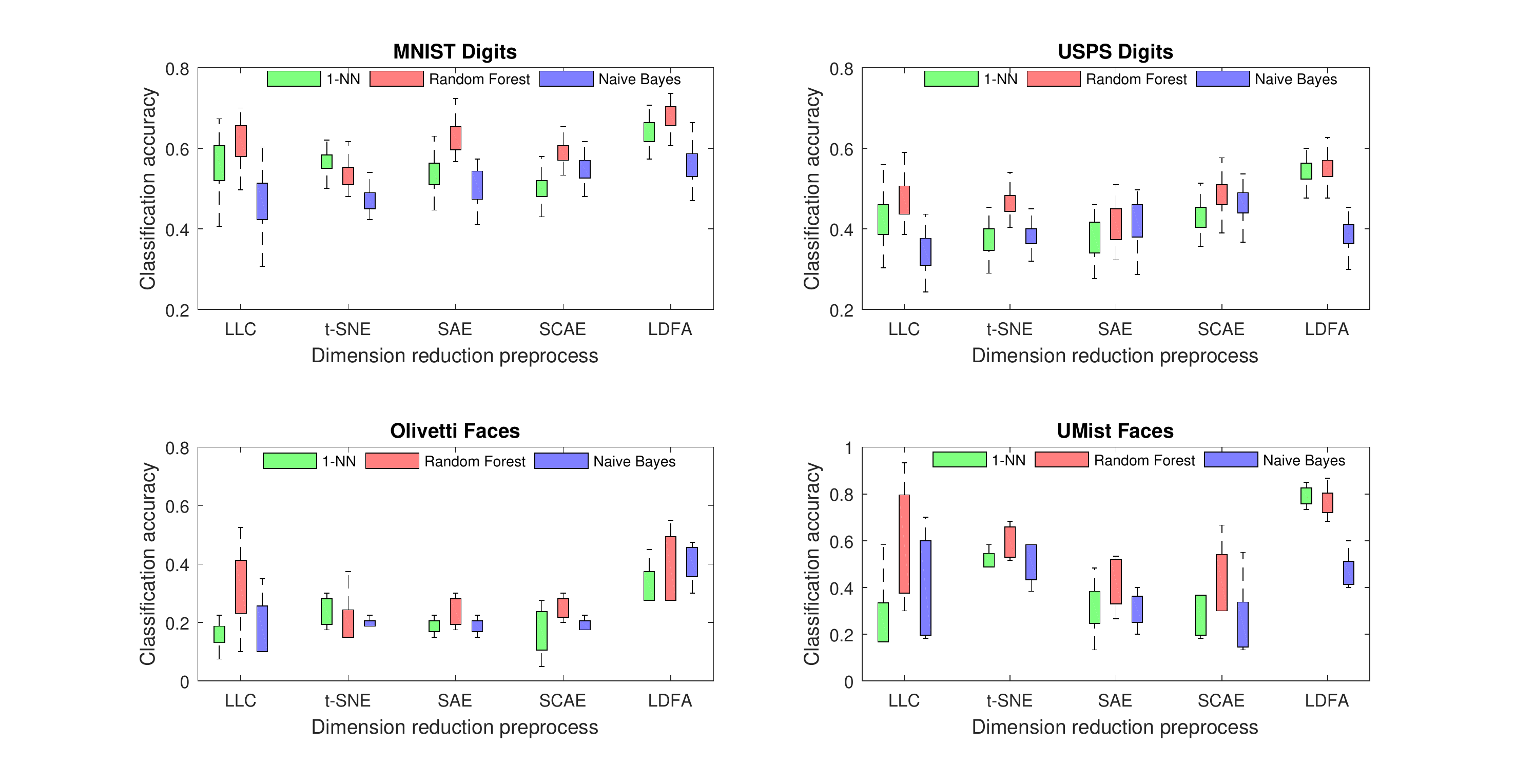}
\caption{Classification accuracies of three common
classification approaches with different dimension reduction
preprocesses on the MNIST Digits, USPS Digits, Olivetti Faces and
UMist Faces data sets.} \label{fig:fourclassification}
\end{figure*}

Guided by similar exploration processes, we
determine the network structures to be applied on the USPS, Olivetti
Faces and UMist Faces data sets as 256-64-30, 4096-64 and 10304-64
respectively. Then we perform dimension reduction again using the
LLC, t-SNE, SAE, SCAE and LDFA, and feed the low-dimensional feature
representations to the 1-NN, random forest \cite{ho1995random} and
the naive Bayes \cite{elkan1997boosting} classifiers. For different
dimension reduction methods, we reduce the data to the same
dimension. The experiment is repeated 10 times with randomly chosen
samples, and the classification accuracies can be found in Fig.
\ref{fig:fourclassification} where the low-dimensional feature
representations learned by the LDFA algorithm achieve the highest
classification accuracy in most cases except for the Bayes
classification on the USPS Digits and UMist Faces data sets. We
believe the good performance of LDFA stems from its ability to learn
not only the global characteristic but also the local deep-level
characteristic of the data sets.


\begin{figure}
\centering
\includegraphics[width = 3.5in]{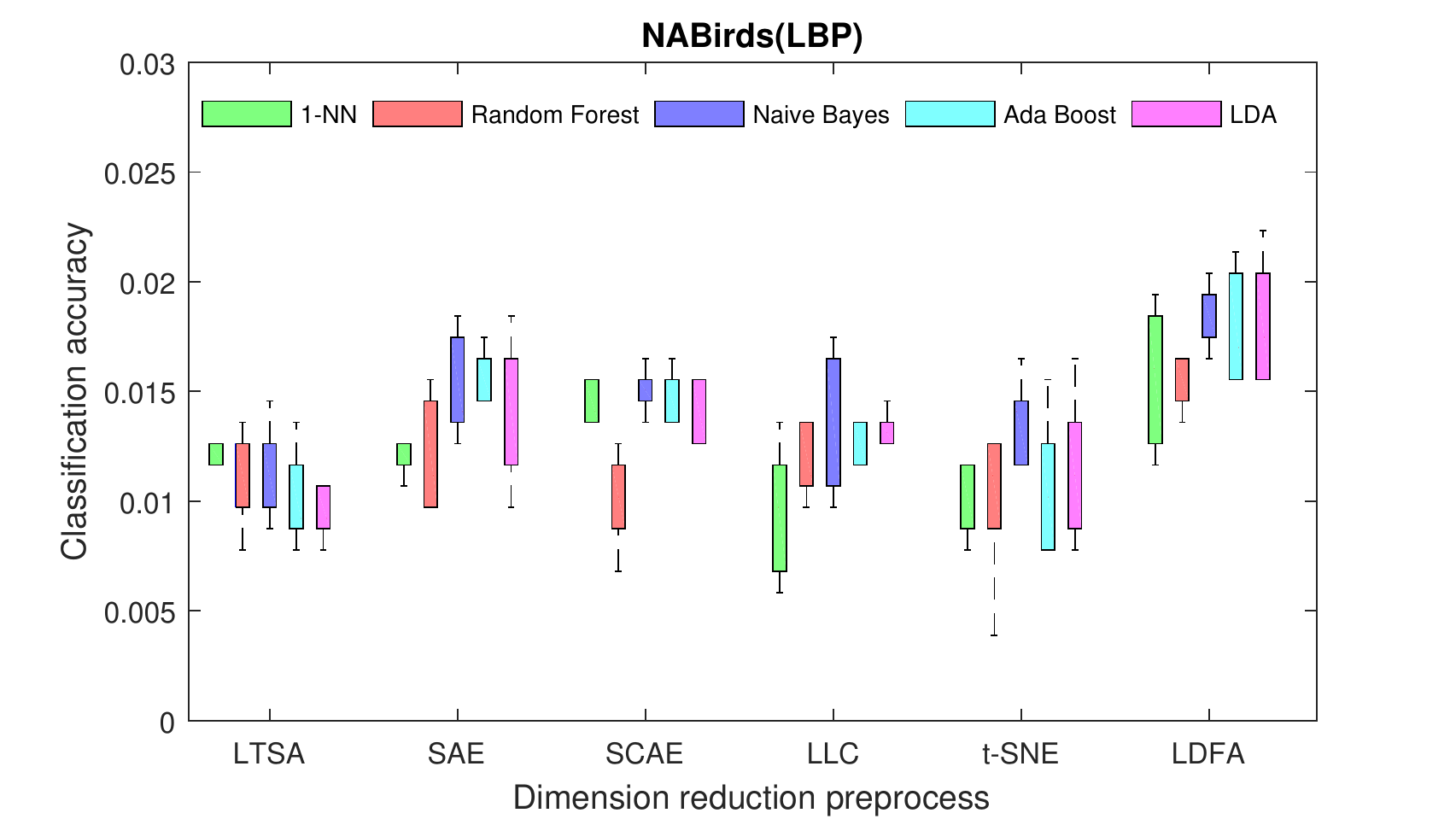}
\caption{Classification accuracies of five common
classification approaches with different dimension reduction
preprocesses on the LBP features of the NABirds data set.}
\label{fig:birdclassification}
\end{figure}

\begin{figure}
\centering
\includegraphics[width = 3.5in]{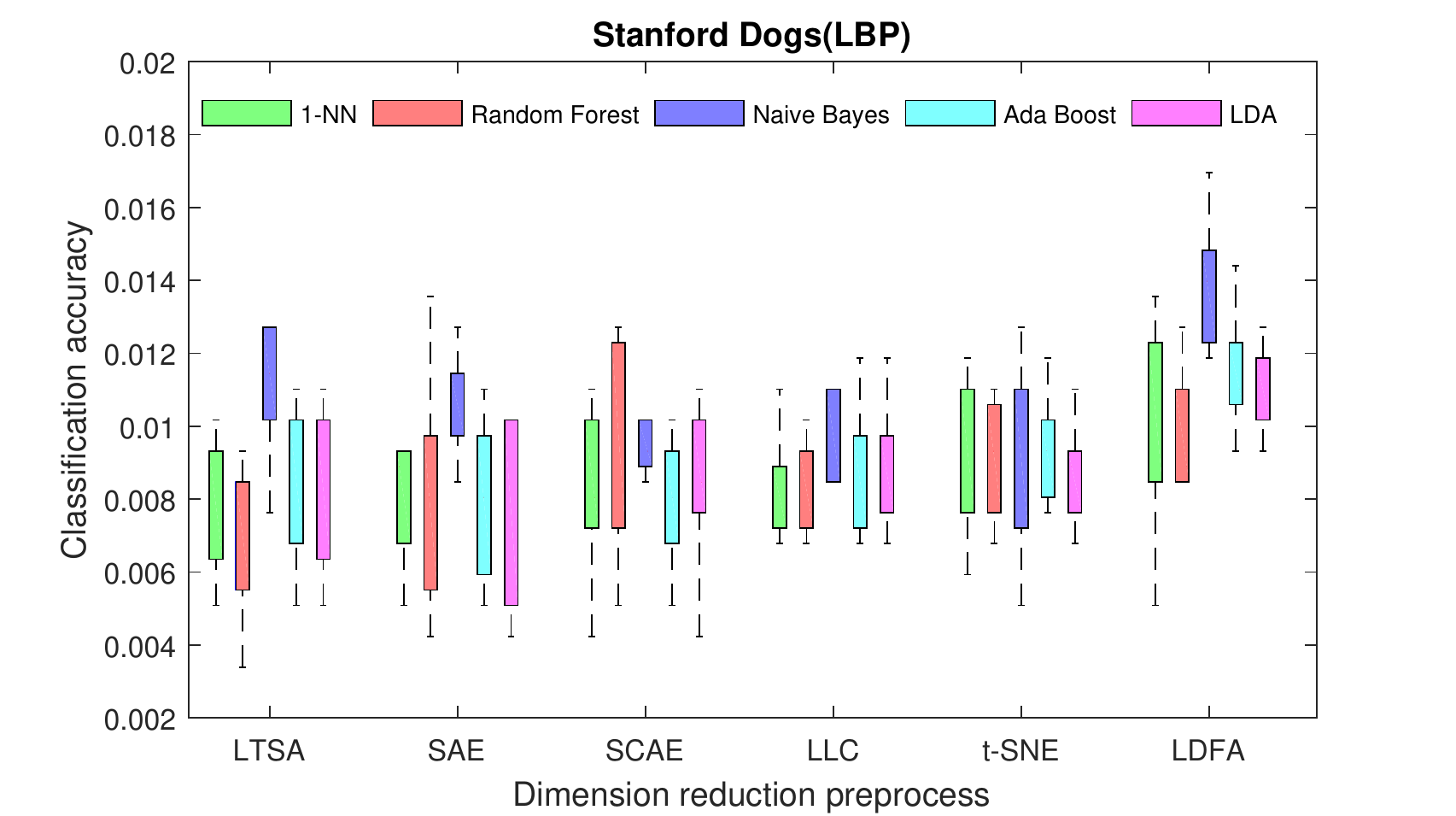}
\caption{Classification accuracies of five common
classification approaches with different dimension reduction
preprocesses on the the LBP features of the Stanford Dogs data
set.} \label{fig:dogclassification}
\end{figure}

\begin{figure}
\centering
\includegraphics[width = 3.5in]{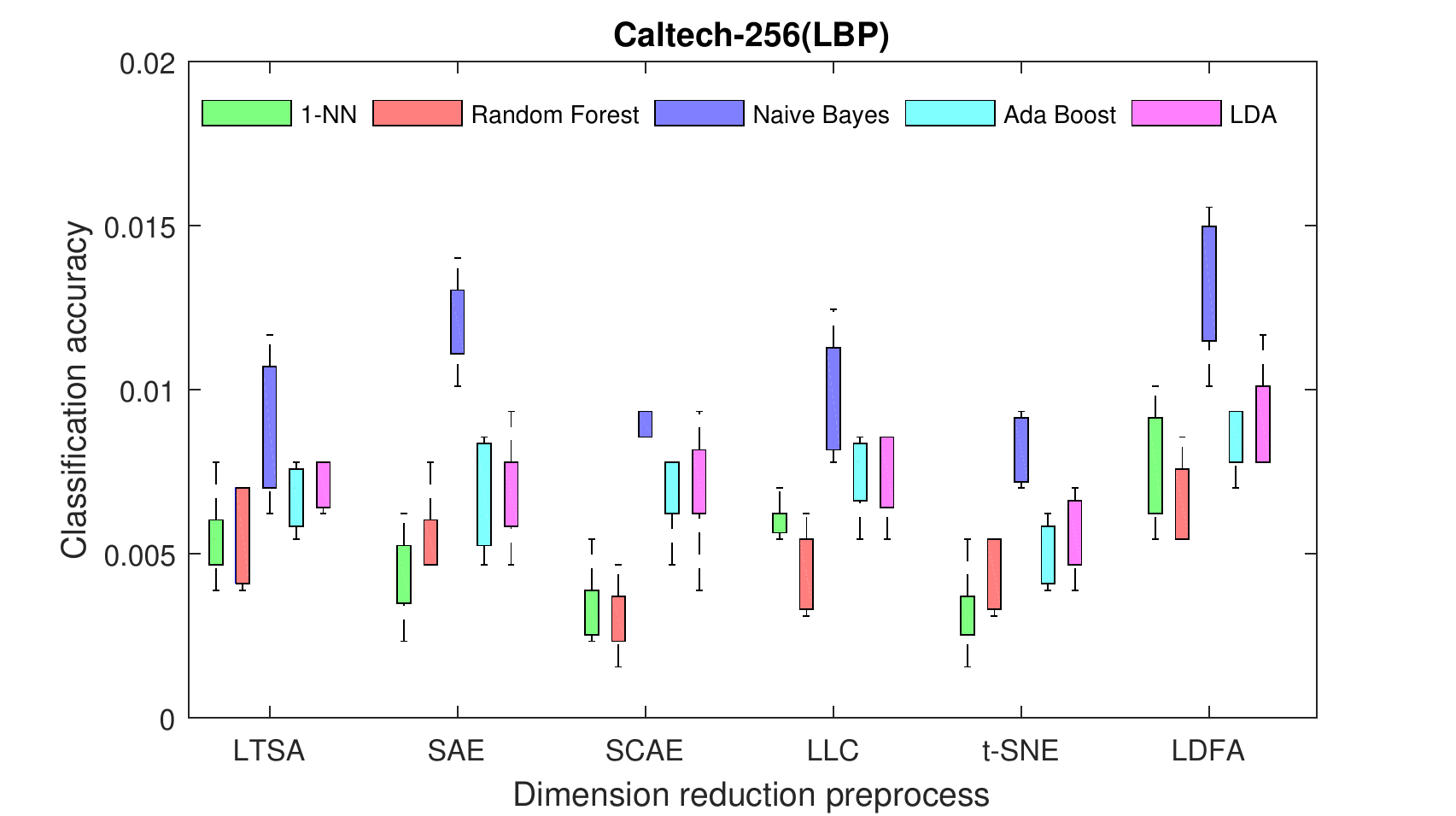}
\caption{Classification accuracies of five common
classification approaches with different dimension reduction
preprocesses on the the LBP features of the Caltech-256 data set.}
\label{fig:objectclassification}
\end{figure}

In addition, we extract the LBP feature descriptors
from the NABirds, Stanford Dogs and Caltech-256 data sets and
conduct classification using the low-dimensional representations
learned from these LBP descriptors by the LTSA, SAE, SCAE, LLC,
t-SNE and the LDFA. We use the same network structure that has been
adopted in Section \ref{visualization}, and this procedure is also
repeated 10 times. The classification results of the 1-NN, random
forest, naive Bayes, AdaBoost ensemble \cite{shen2006ensemble}, and
LDA \cite{fisher1936use} classifiers are shown in Fig.
\ref{fig:birdclassification} through Fig.
\ref{fig:objectclassification}, which indicate that the LDFA
produces the best classification results in that the features
learned by LDFA generate not only the highest but also the best mean
classification accuracy with respect to each classification
algorithm. The only exception is the random forest classification on
the Stanford Dogs data set.

\begin{figure}
\centering
\includegraphics[width = 3.5in]{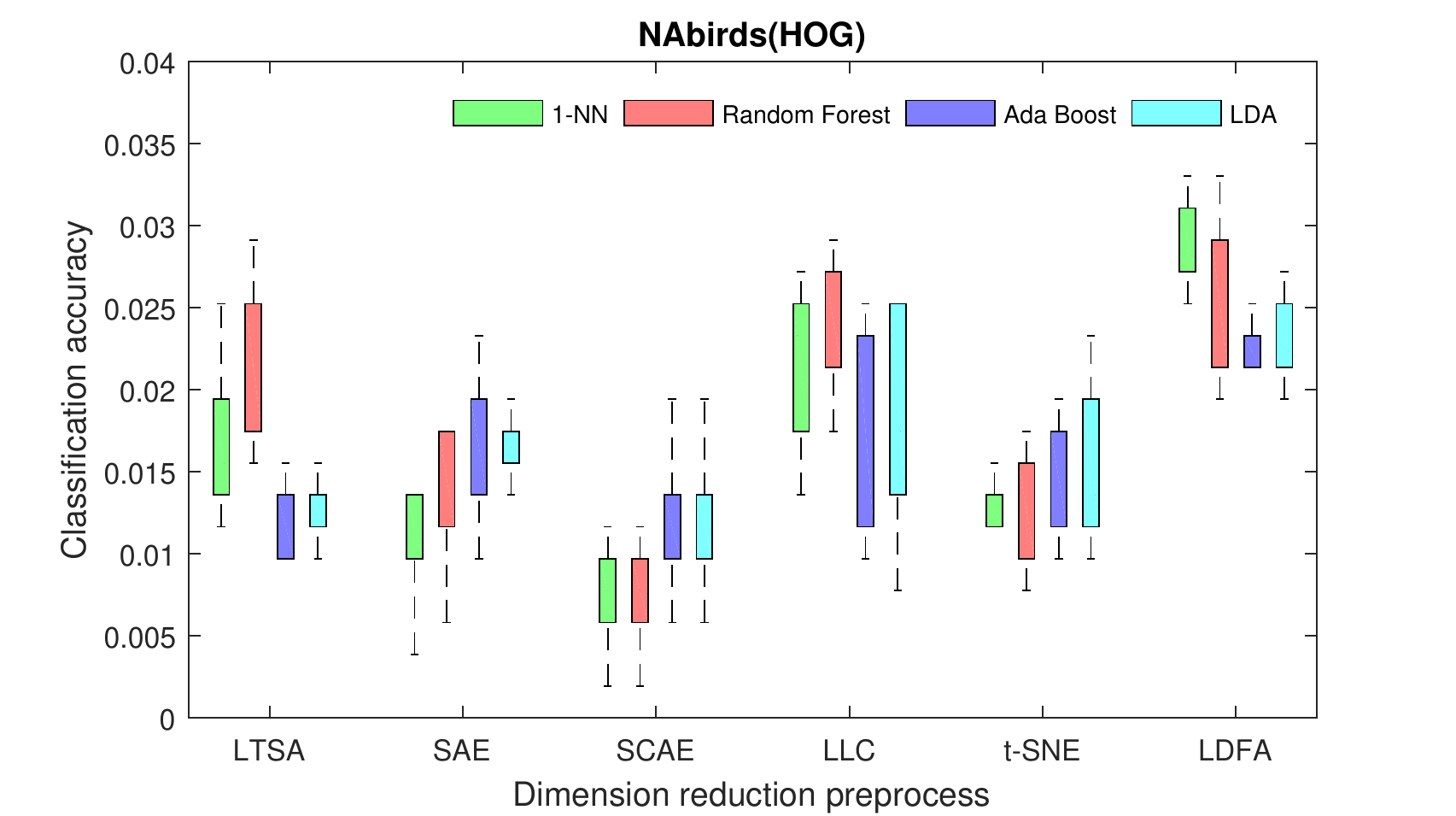}
\caption{Classification accuracies of four common
classification approaches with different dimension reduction
preprocesses on the HOG features of the NABirds data set.}
\label{fig:birdclassificationhog}
\end{figure}

\begin{figure}[ht]
\centering
\includegraphics[width = 3.5in]{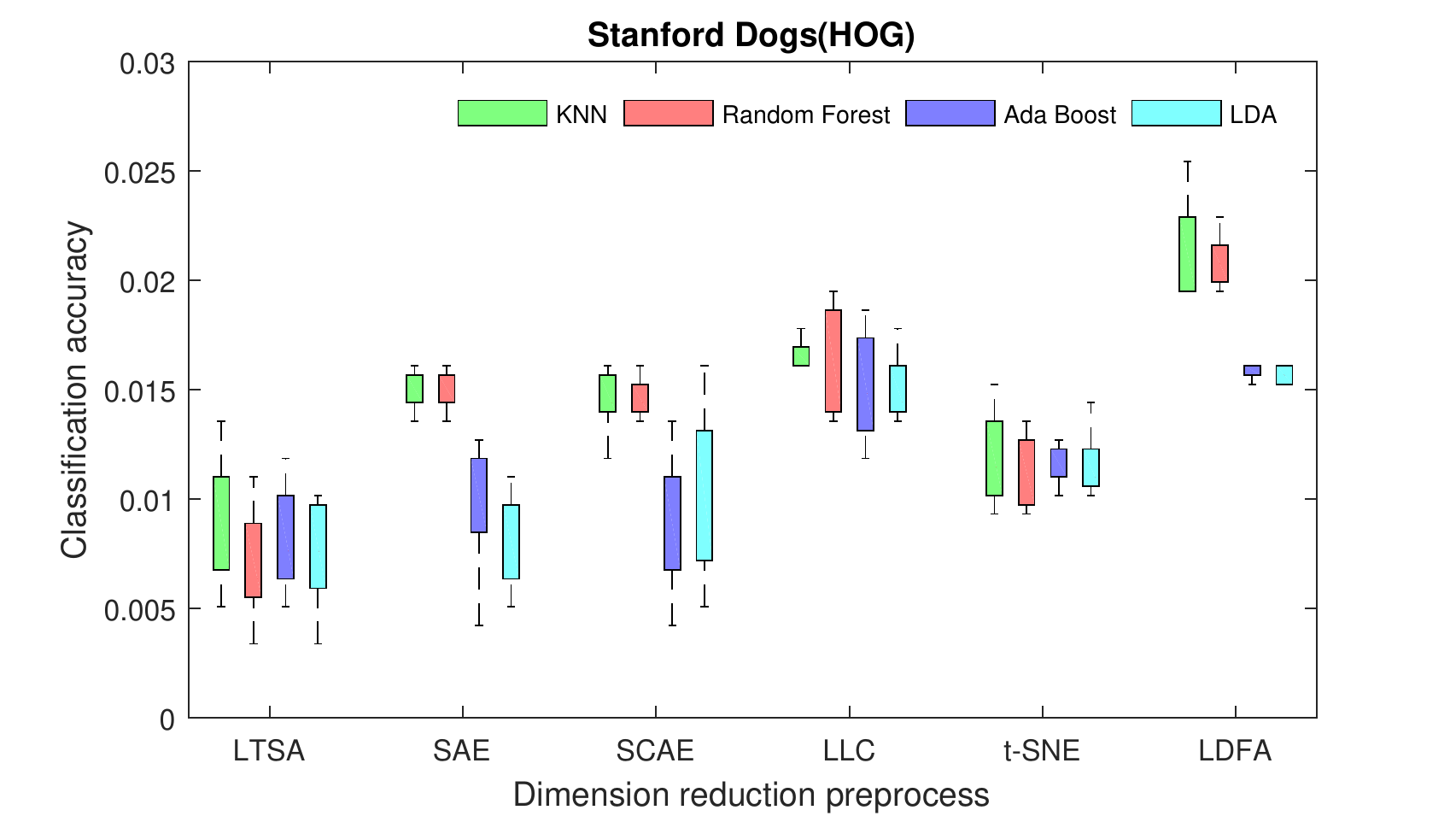}
\caption{Classification accuracies of four common
classification approaches with different dimension reduction
preprocesses on the HOG features of the Stanford Dogs data set.}
\label{fig:dogclassificationhog}
\end{figure}

\begin{figure}
\centering
\includegraphics[width = 3.5in]{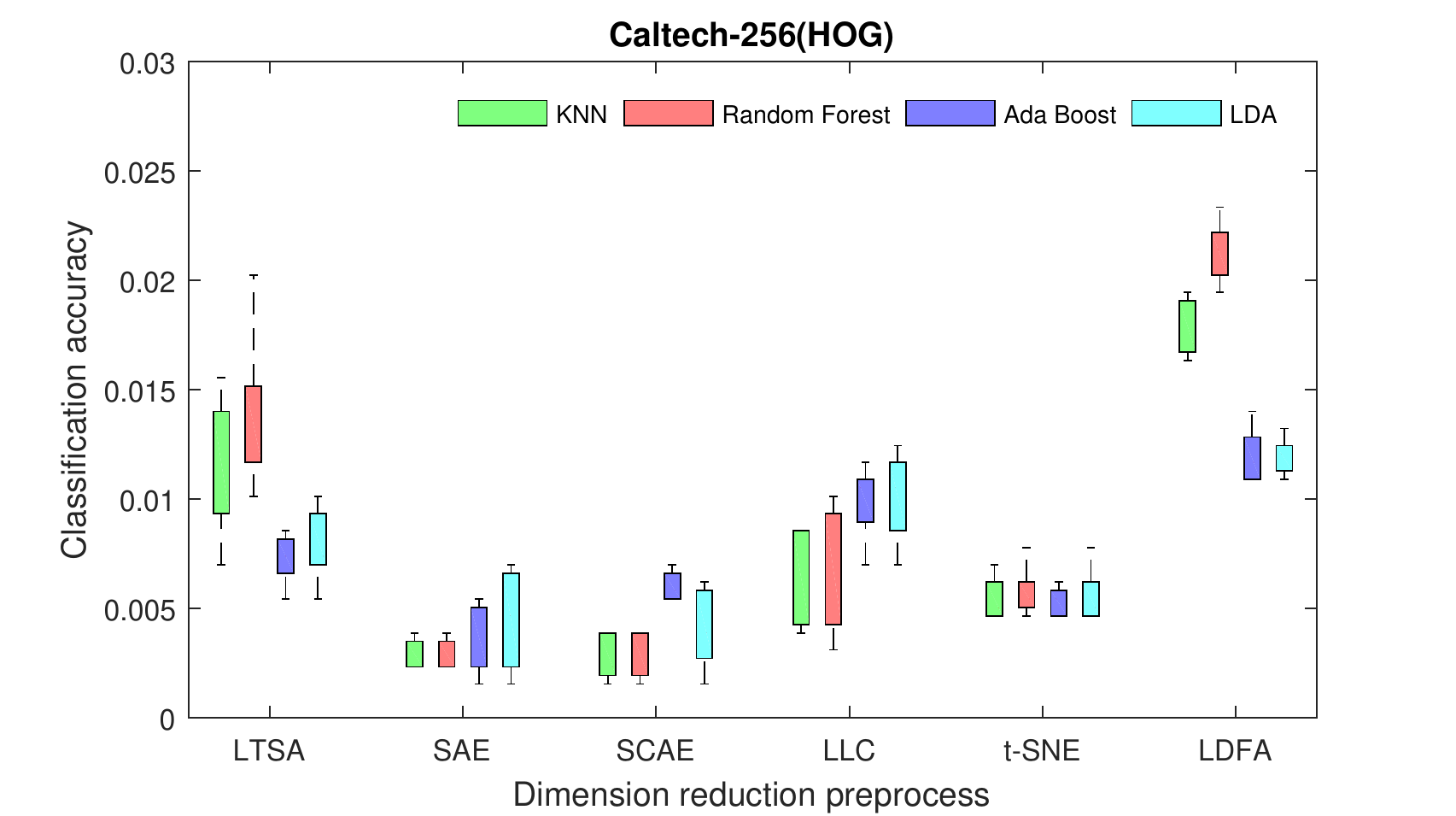}
\caption{Classification accuracies of four common
classification approaches with different dimension reduction
preprocesses on the HOG features of the Caltech-256 data set.}
\label{fig:objectclassificationhog}
\end{figure}

Furthermore, we use Histogram of Oriented Gradients
(HOG) \cite{dalal2005histograms} feature descriptors of the NABirds,
Stanford Dogs and Caltech-256 data sets to evaluate the
aforementioned dimension reduction methods. Specifically, we
re-scale all the bird images to the same size of 512$\times$512, and
set the size of the cells as 8$\times$8. Hence, the extracted HOG
descriptors are of 142884 dimensions. Similarly, we resize all the
dog and object images to 256$\times$256 while keeping the cell size
unchanged to obtain 34596-dimension HOG descriptors. For the sake of
computational convenience, the dimensions of all the HOG descriptors
are reduced to 500 through PCA firstly. Then we use the LTSA, SAE,
SCAE, LLC, t-SNE and LDFA to extract 30-dimension representations
from the 500-dimension HOG descriptors, and conduct the
classification based on the extracted representations. The
experiment is also performed 10 times with randomly chosen data each
time. The classification accuracies of four classification
algorithms on the NABirds, Stanford Dogs and Caltech-256 data sets
are demonstrated in Fig. \ref{fig:birdclassificationhog} through
Fig. \ref{fig:objectclassificationhog} where the LDFA still produces
the best results except for the AdaBoost and LDA classification on
Stanford Dogs data set. Therefore, we believe the LDFA algorithm can
learn more discriminative and more robust feature representations.

\section{Conclusion}
We have proposed an unsupervised deep-learning method named Local
Deep-Feature Alignment (LDFA). We define a neighbourhood for each data
sample and learn the local deep features via SCAEs. Then we align
the local features with global features by local affine
transformations. Additionally, we provide an explicit approach to mapping new
data into the learned low-dimensional subspace.

The proposed LDFA method has been used as a pre-processing step for
image visualization, image clustering and image classification in
our experiments. We found that SCAE could extract discriminative
local deep features robustly from a small number of data samples
(neighbourhood) with few network layers. These experimental results
persuaded us that using SCAE to capture the local characteristics of
data sets would improve the performance of the unsupervised
deep-learning method.


\ifCLASSOPTIONcaptionsoff
  \newpage
\fi

\bibliographystyle{IEEEtran}
\bibliography{Local-Deep-Feature-Alignment}

%
\begin{IEEEbiography}[{\includegraphics[width=1in,height=1.25in,clip,keepaspectratio]{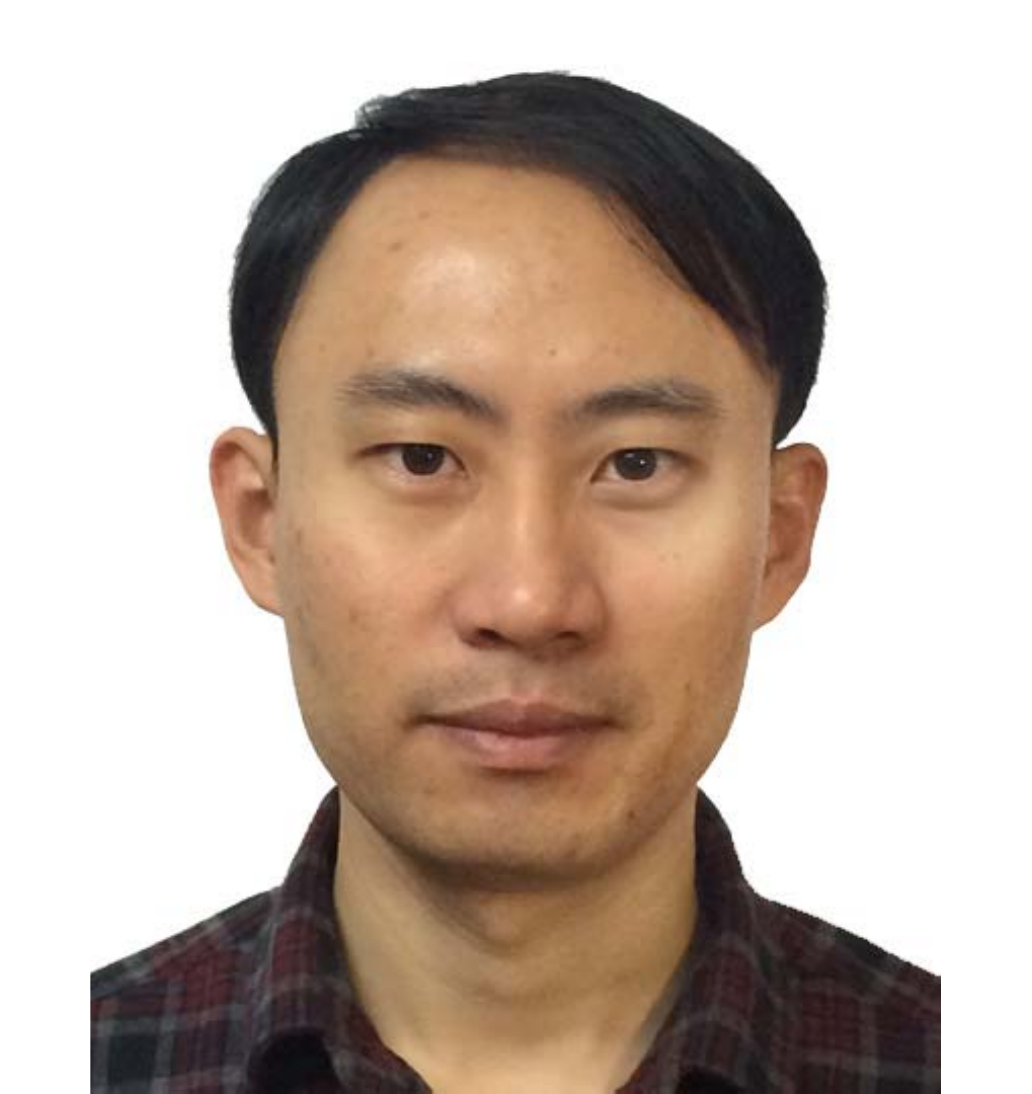}}]{Jian Zhang} received the Ph.D. degree from Zhejiang University, Zhejiang, China. He is currently an Associate Professor with the School of Science and Technology, Zhejiang International Studies University, Hangzhou, China. From 2009 to 2011, he was with Department of Mathematics of Zhejiang university as a Post-doctoral Research Fellow. In 2016, he had been doing research on machine learning at Simon Fraser University (SFU) as a Visiting Scholar. His research interests include but not limited to machine learning, computer animation and image processing. He serves as a reviewer of several prestigious journals in his research domain.

\end{IEEEbiography}


\begin{IEEEbiography}[{\includegraphics[width=1in,height=1.25in,clip,keepaspectratio]{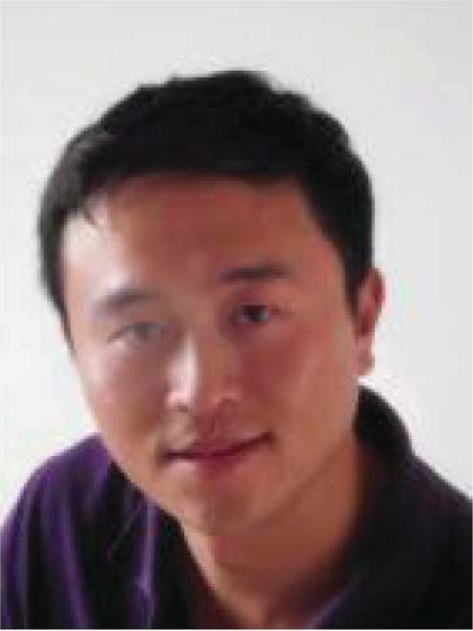}}]{Jun Yu} (M'13) received the B.Eng. and Ph.D. degrees from Zhejiang University, Zhejiang, China.
He is currently a Professor with the School of Computer Science and Technology, Hangzhou Dianzi University, Hangzhou, China. He was an Associate Professor with the School of Information Science and Technology, Xiamen University, Xiamen, China. From 2009 to 2011, he was with Nanyang Technological University, Singapore. From 2012 to 2013, he was a Visiting Researcher at Microsoft Research Asia (MSRA). Over the past years, his research interests have included multimedia analysis, machine learning, and image processing. He has authored or coauthored more than 60 scientific articles. In 2017 Prof. Yu received the IEEE SPS Best Paper Award.
Prof. Yu has (co-)chaired several special sessions, invited sessions, and workshops. He served as a program committee member or reviewer of top conferences and prestigious journals. He is a Professional Member of the Association for Computing Machinery (ACM) and the China Computer Federation (CCF).
\end{IEEEbiography}


\begin{IEEEbiography}[{\includegraphics[width=1in,height=1.25in,clip,keepaspectratio]{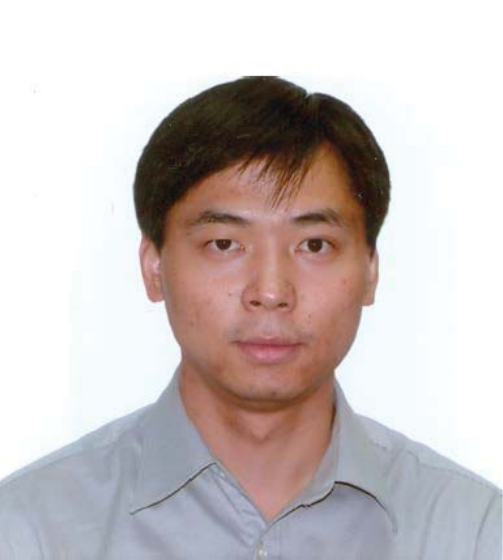}}]{Dacheng Tao} (F¡¯15) is Professor of Computer Science and ARC Laureate Fellow in the School of Information Technologies and the Faculty of Engineering and Information Technologies, and the Inaugural Director of the UBTECH Sydney Artificial Intelligence Centre, at the University of Sydney. He mainly applies statistics and mathematics to Artificial Intelligence and Data Science. His research interests spread across computer vision, data science, image processing, machine learning, and video surveillance. His research results have expounded in one monograph and 500+ publications at prestigious journals and prominent conferences, such as IEEE T-PAMI, T-NNLS, T-IP, JMLR, IJCV, NIPS, ICML, CVPR, ICCV, ECCV, ICDM; and ACM SIGKDD, with several best paper awards, such as the best theory/algorithm paper runner up award in IEEE ICDM¡¯07, the best student paper award in IEEE ICDM¡¯13, the distinguished student paper award in the 2017 IJCAI, the 2014 ICDM 10-year highest-impact paper award, and the 2017 IEEE Signal Processing Society Best Paper Award. He received the 2015 Australian Scopus-Eureka Prize, the 2015 ACS Gold Disruptor Award and the 2015 UTS Vice-Chancellor¡¯s Medal for Exceptional Research. He is a Fellow of the IEEE, AAAS, OSA, IAPR and SPIE.
\end{IEEEbiography}

\end{document}